\pdfoutput=1

\documentclass[11pt]{article}

\usepackage[final]{acl}

\usepackage{times}
\usepackage{latexsym}

\usepackage[T1]{fontenc}

\usepackage[utf8]{inputenc}

\usepackage{microtype}

\usepackage{booktabs}
\usepackage{amsmath}
\usepackage{amssymb}
\usepackage{algorithm}
\usepackage{algpseudocode}
\usepackage{multirow}
\usepackage{mathtools}
\usepackage{subfigure}
\usepackage{graphicx}
\usepackage{balance}
\usepackage{color}
\usepackage{xcolor}
\usepackage{enumitem}
\usepackage{xspace}

\newcommand{\stitle}[1]{\vspace{1mm} \noindent {\bf #1}}
\newcommand{\eg}{{\it e.g.}}

\newcommand{\ie}{{\it i.e.}}

\newcommand{\model}{SIBO}

\usepackage{inconsolata}

\title{SIBO: A Simple Booster for Parameter-Efficient Fine-Tuning}

\author{
  Zhihao Wen$^{1*}$ \quad Jie Zhang$^{2}$\thanks{ Co-first authors with equal contribution.  $\dagger$ Corresponding author.}  \quad Yuan Fang$^{1\dagger}$   \\
  $^1$Singapore Management University \\
  $^2$National University of Singapore \\
  \texttt{zhwen.2019@smu.edu.sg, jiezhang\_jz@u.nus.edu, yfang@smu.edu.sg} \\
}

\begin{document}
\maketitle

\begin{abstract}

Fine-tuning all parameters of large language models (LLMs) necessitates substantial computational power and extended time. 
Latest advancements in parameter-efficient fine-tuning (PEFT) techniques, such as Adapter tuning and LoRA, allow for adjustments to only a minor fraction of the parameters of these LLMs.
Concurrently, it has been noted that the issue of over-smoothing diminishes the effectiveness of these Transformer-based LLMs, resulting in suboptimal performances in downstream tasks.
In this paper, we present  \model, which is a \underline{SI}mple \underline{BO}oster to enhance PEFT, by injecting an \emph{initial residual}.
\model~is straightforward and readily extensible to a range of state-of-the-art PEFT techniques to alleviate over-smoothing and enhance performance. Extensive experiments on 22 benchmark datasets demonstrate that \model\ 
significantly enhances the performance of various strong baselines, achieving up to 15.7\% and 23.5\% improvement over existing PEFT methods on the arithmetic and commonsense reasoning tasks, respectively.
\end{abstract}

\section{Introduction}
Many Transformer-based large language models (LLMs) exhibit significant depth, \eg,  BERT-large \cite{devlin2018bert} has 24 layers, LLaMA-7B \cite{touvron2023llama} has 32 layers, and LLaMA-65B has 80 layers. Yet, this depth presents a challenge \cite{zhou2021deepvit, gong2021vision}: Deep Transformers tend to encounter the \emph{over-smoothing} problem. This issue, as detailed by \citet{brunner2019identifiability}, manifests in the deeper layers of Transformers, where token representations increasingly converge toward uniformity. The over-smoothness not only impedes the scalability of Transformer training, particularly in terms of depth, but also limits the efficacy of scaling up the model size. Consequently, expanding the model often results in marginal enhancements or, in some cases, reduced accuracy \cite{xue2023study}.

Meanwhile, a significant drawback of  full-model fine-tuning for LLMs is that 
it requires updating all the parameters of the original model.
While this constitutes a relatively minor limitation for models like BERT-large \cite{devlin2018bert} or RoBERTa-large \cite{liu2019roberta}, it escalates into a major obstacle for larger models such as LLaMA \cite{touvron2023llama}, which contain billions of trainable parameters. 
Many approaches \cite{houlsby2019parameter, hu2021lora, li2021prefix, lester2021power} have been explored to address this issue by updating only a subset of the parameters or lightweight external modules tailored for new tasks. Such strategies require storing and loading a relatively small number of task-specific parameters alongside the pre-trained model for each task. These compelling alternatives to full-model fine-tuning are called \emph{parameter-efficient fine-tuning} (PEFT) \cite{houlsby2019parameter}, which significantly enhances the feasibility of deploying LLMs.

Although some approaches have been proposed to deal with the over-smoothing problem, such as adding specifically designed regularization to avoid ``uniform tokens'' \cite{zhou2021deepvit, gong2021vision} and fusing the representations from all layers \cite{shi2022revisiting}, no PEFT method has yet been proposed to alleviate the over-smoothing issue. In the era of LLMs, when internal modifications to models are infeasible for most use cases, addressing the over-smoothing issue through PEFT techniques becomes critical.

\stitle{Challenge and present work.}
Given that existing solutions to over-smoothing involve changes to the model architecture and are hence not parameter-efficient, the question arises: How can over-smoothing be effectively addressed for PEFT techniques? 
Two primary factors may contribute to the over-smoothing problem: 1) \emph{redundancy} within the model's encoding layers, and 2) a \emph{suboptimal training process} that hinders the effective optimization of the deeper layers. To address the first issue, a straightforward and logical solution is to reduce the number of layers in the encoder. However, this approach can result in a decline in performance \cite{chen2023alleviating}.
To address the second issue, previous approaches \cite{gong2021vision,zhou2021deepvit,shi2022revisiting}
are not parameter-efficient, thereby limiting their application to LLMs.

To devise a flexible yet simple plug-and-play framework for alleviating over-smoothing with existing PEFT techniques, 
our idea boils down to injecting an initial residual into the PEFT input. 
This initial residual connection ensures that the final representation of each token preserves at least a minimum portion of the input layer's features, aiming to reduce the uniformity of the final token representations. We name the novel framework \model, a \underline{SI}mple \underline{BO}oster to enhance PEFT techniques designed for LLMs, most notably Adapter  \cite{houlsby2019parameter} and LoRA \cite{hu2021lora}. 

Empirically, on the arithmetic reasoning task, 
\model\ outperforms Adapter and LoRA by up to 15.7\% and 13.6\%, respectively. On the commonsense reasoning task, the improvement is up to 7.6\% over Adapter, and 23.5\% over LoRA. 

\section{Preliminaries}
In the following, we present a summary of two popular lines of PEFT techniques: 
adapters and reparametrization-based
methods.

\stitle{Adapters.}
Adapters fall into two distinct categories: parallel and serial adapters. Parallel adapters \cite{he2021towards} integrate additional learnable modules alongside various layers of the core model. In contrast, series adapters \cite{houlsby2019parameter} insert these modules sequentially between specific layers, \eg, adding fully connected networks after both the attention and feed forward layers in the Transformer model. 
In this work, we focus on the classical serial adapter, which has the following general formulation:
\begin{align}
    \mathbf{h} \leftarrow \mathbf{h} + f(\mathbf{h}\mathbf{W}_{\text{down}})\mathbf{W}_{\text{up}},
    \label{eq:adapter}
\end{align}
where $\mathbf{h}\in \mathbb{R}^{1\times d}$ represents the output of the preceding layer, after which the adapter is inserted.  Consequently, $\mathbf{h}$ serves as the input to the adapter. It first undergoes a down-projection to a lower dimension $r$ via $\mathbf{W}_{\text{down}} \in \mathbb{R}^{d \times r}$, followed by an up-projection back to its original dimension $d$ via $ \mathbf{W}_{\text{up}} \in \mathbb{R}^{r \times d} $. The function $f(\cdot)$ represents a non-linear function.

\stitle{Reparameterization-based methods.}
These methods are designed to modify network weights through a low-rank strategy. This technique effectively reduces the number of tunable parameters without compromising performance. 
For example, Low-Rank Adaptation (LoRA) approximates the update $\Delta \mathbf{W}$ to a pre-trained weight matrix $\mathbf{W} \in \mathbb{R}^{d \times d}$ through a low-rank decomposition:
\begin{align}
    \mathbf{h} \leftarrow \mathbf{h} (\mathbf{W} + s \cdot \mathbf{W}_{\text{down}}\mathbf{W}_{\text{up}}),
    \label{eq:lora}
\end{align}
where $\mathbf{h} \in \mathbb{R}^{d}$ is the output of the  preceding layer, and $\mathbf{W} \in \mathbb{R}^{d \times d} $ is a pre-trained weight matrix, \eg, for multilayer perceptron (MLP) or attention layers. The matrices $\mathbf{W}_{\text{down}} \in \mathbb{R}^{d \times r}$ and $\mathbf{W}_{\text{up}} \in \mathbb{R}^{r \times d}$ are lower-rank matrices to approximate the update, \ie, $\Delta \mathbf{W} \approx \mathbf{W}_{\text{down}}\mathbf{W}_{\text{up}}$. Here, $r \ll d$ serves as a crucial hyperparameter for LoRA, while the scalar $ s \geq 1$ is an adjustable hyperparameter.

\section{Methodology}
In this section, we first analyze the over-smoothing issue in PEFT techniques, and subsequently present our proposed framework, \model.

\subsection{Over-smoothing in PEFT}
Originating from graph neural networks, the term \emph{over-smoothing} denotes a decline in performance attributed to the increasing homogeneity of node representations \cite{li2018deeper, xu2018representation, huang2020tackling},  stemming from the repetitive use of the same adjacency matrix in successive aggregation layers.
\citet{shi2022revisiting} have since identified an over-smoothing phenomenon in language models as well, wherein distinct tokens in an input sentence exhibit increasingly similar representations as more layers are stacked,
diminishing the effectiveness of deep Transformer models.

While several strategies have been proposed to mitigate over-smoothing \cite{zhou2021deepvit, gong2021vision,shi2022revisiting}, they are not designed for PEFT techniques, making them less practical for LLMs.
In particular, we also observe over-smoothing in widely adopted PEFT techniques including adapters and LoRA, especially with deep layers, through quantitative analysis. 
In our analysis, over-smoothing can be detected by assessing the similarity among tokens in the same sentence, known as \emph{token-wise cosine similarity}. Given a sentence consisting of $m$ tokens, represented by $( \mathbf{h}_1, \mathbf{h}_2, \ldots, \mathbf{h}_m )$, its token-wise cosine similarity is computed as
\begin{align}
    \frac{1}{m(m-1)} \sum_{i \neq j} \frac{\mathbf{h}_i^\top \mathbf{h}_j}{\|\mathbf{h}_i\|_2 \|\mathbf{h}_j\|_2},
    \label{eq:tokenwise_sim}
\end{align}
where $\|\cdot\|_2$ is the Euclidean norm.
As shown in Figs.~\ref{fig:token sim investigate, bert} and \ref{fig:token sim investigate, llm}, with both adepters and LoRA, an increase in token-wise similarity is noted consistently as the layer depth in the backbone language model increases. Hence, the issue of over-smoothing also persists in pre-trained language models that have undergone adaptation via PEFT techniques.
Therefore, it is imperative to devise a general framework that eases over-smoothing for PEFT methods, while retaining their  efficiency.

\begin{figure}
   \subfigure[CoLA]{
   \centering
   \includegraphics[width=0.47\linewidth]{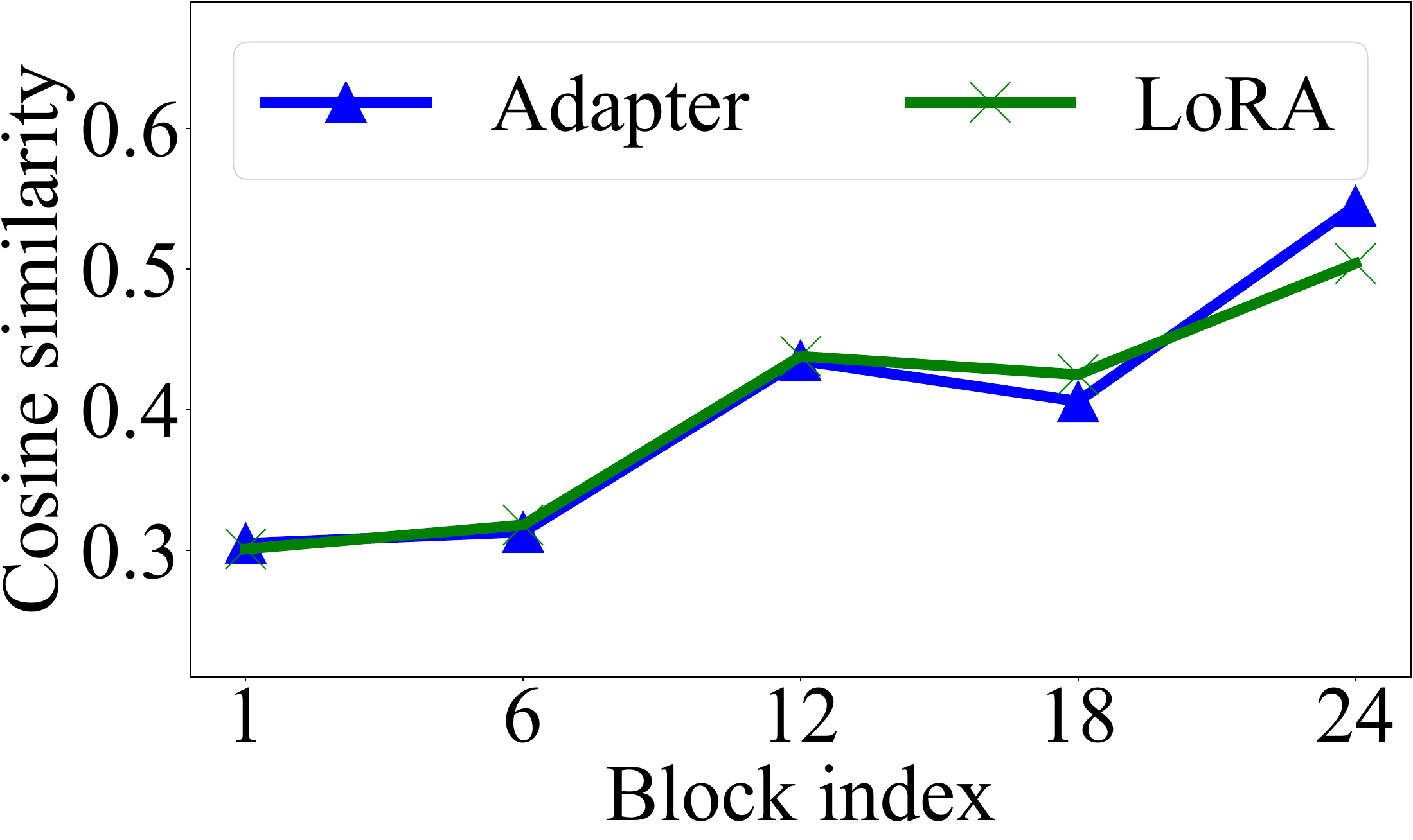}
   }
   \subfigure[STS-B]{
   \centering
   \includegraphics[width=0.47\linewidth]{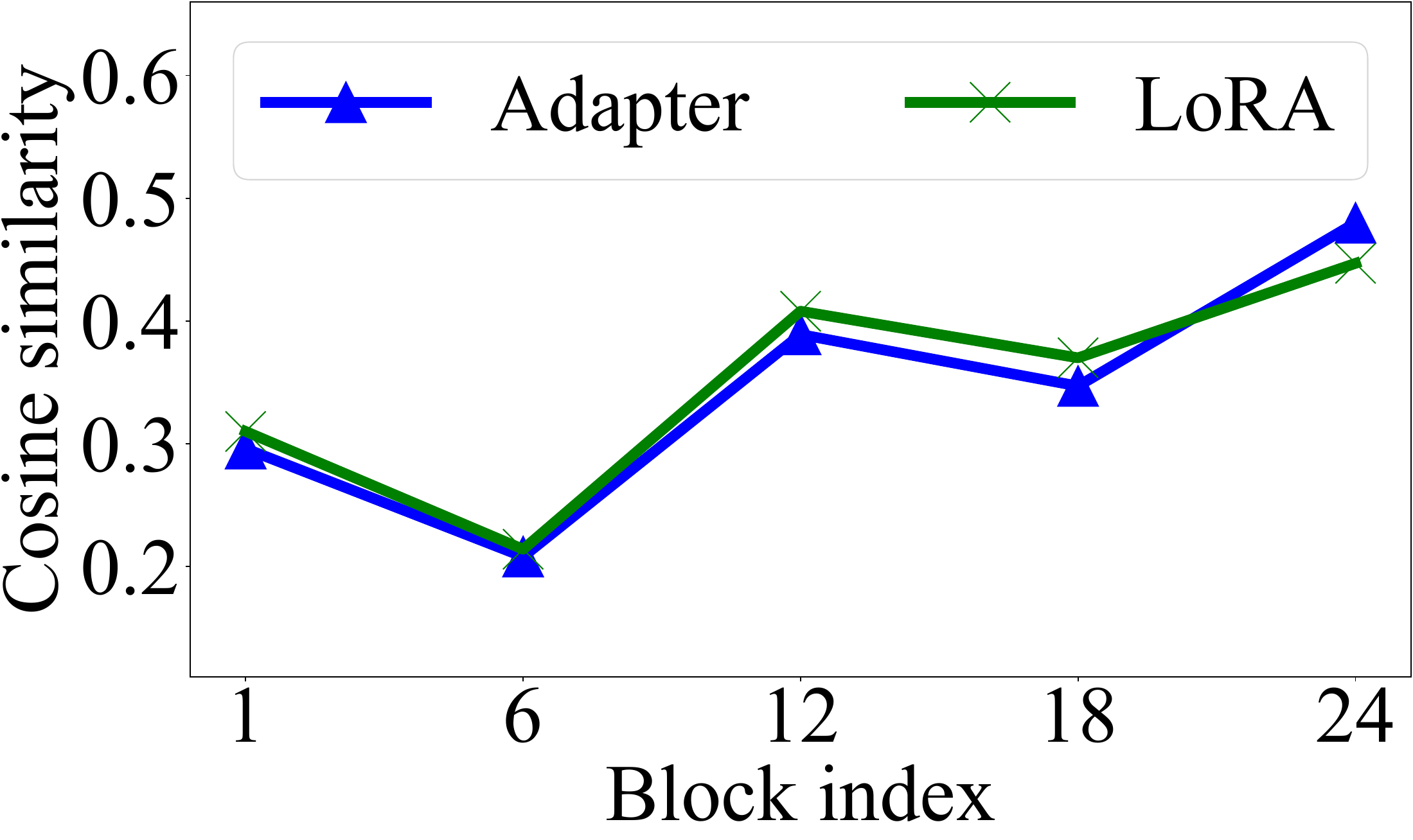}
   }
    \vspace{-2mm}
	\caption{Over-smoothing in PEFT. The results are the averaged token-wise similarity of sentences in the test sets of the corpora in the GLUE benchmark \cite{wang2018glue}, with BERT-large as the backbone.}
	\label{fig:token sim investigate, bert}
\end{figure}

\begin{figure}
   \subfigure[MAWPS]{
   \centering
   \includegraphics[width=0.47\linewidth]{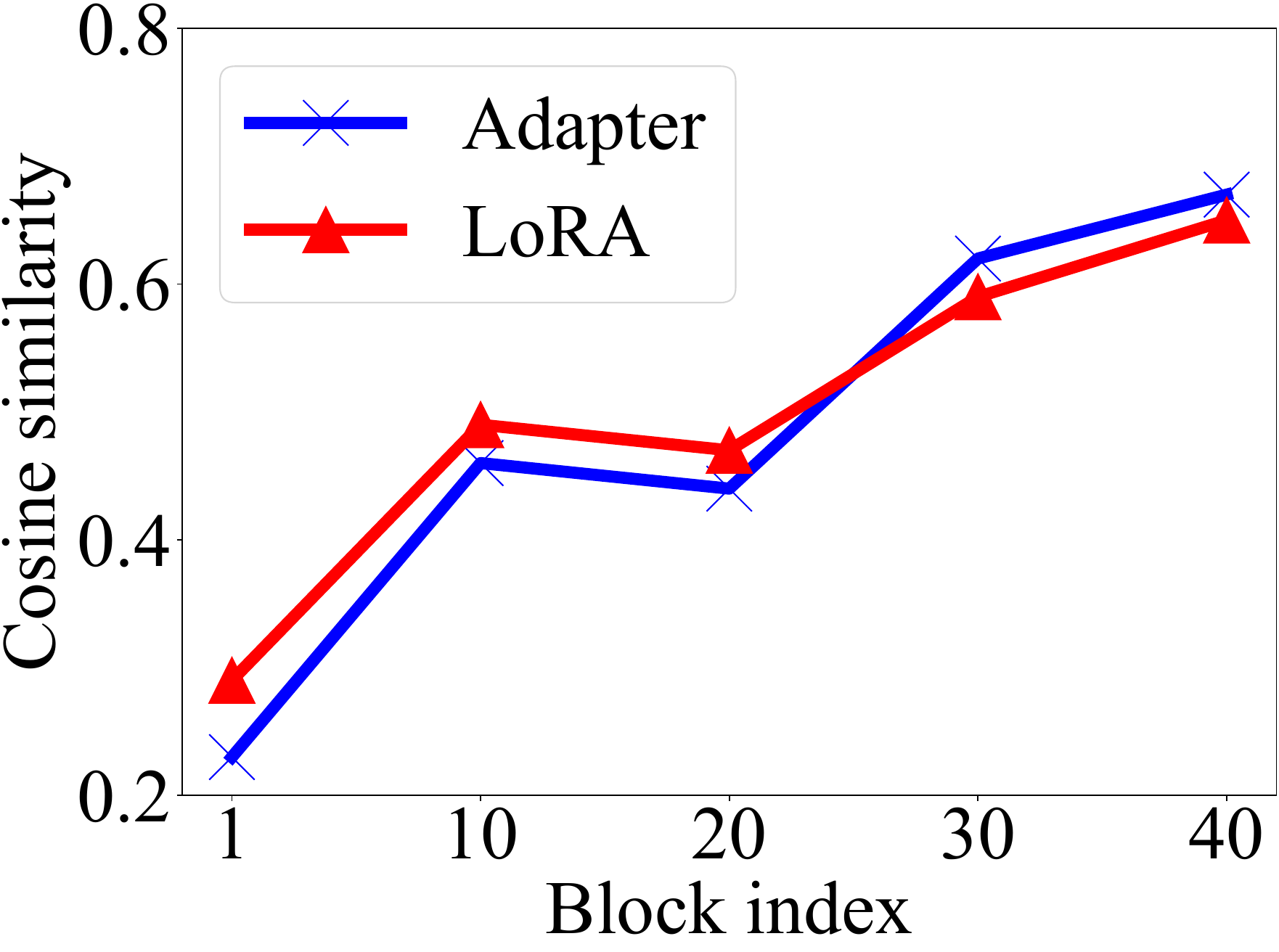}
   }
   \subfigure[SVAMP]{
   \centering
   \includegraphics[width=0.47\linewidth]{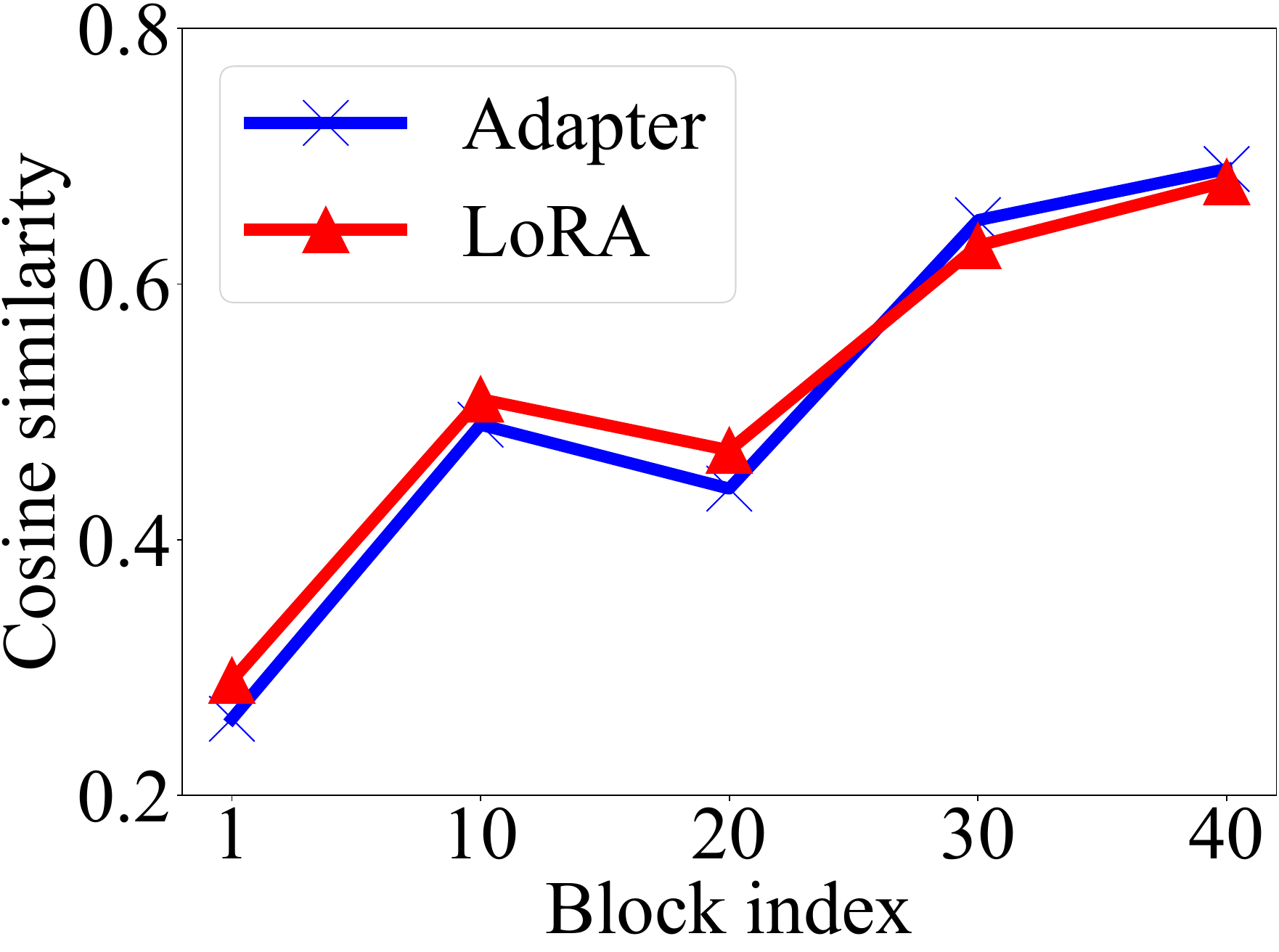}
   }
    \vspace{-2mm}
	\caption{Over-smoothing in PEFT. The results are the averaged token-wise similarity of sentences in the test sets of MAWPS \cite{koncel2016mawps} and SVAMP \cite{patel2021nlp}, with LLaMA (13B) as the backbone.}
	\label{fig:token sim investigate, llm}
\end{figure}

\begin{figure}
   \centering
   \includegraphics[width=0.99\linewidth]{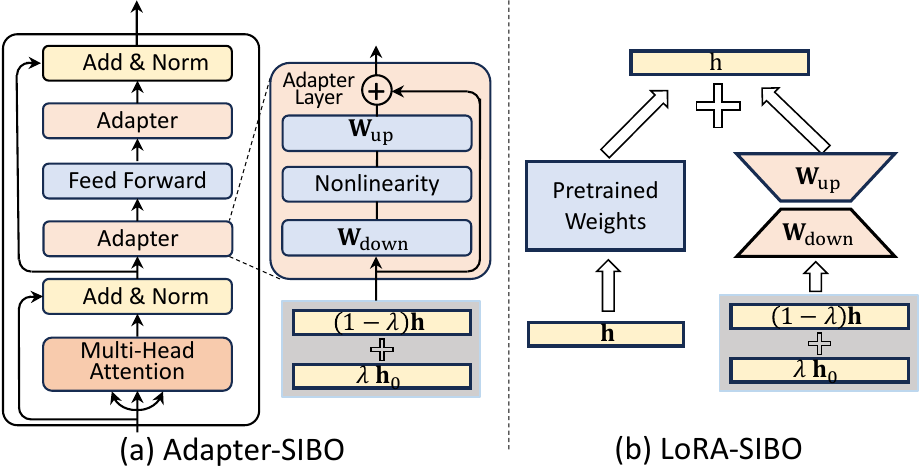}%
    \vspace{-1mm}%
	\caption{Proposed framework \model, applying to two popular PEFT methods: (a) Adapter, and (b) LoRA.}
	\label{fig:frame}
\end{figure}

\subsection{Initial residual integration}
To achieve a universal plug-and-play enhancement for PEFT, we start with the input to the PEFT module, and inject an initial residual into the input at each layer of the pre-trained model. 

Let the initial token representation serving as input to the pre-trained model be denoted by $\mathbf{h}_{0}\in \mathbb{R}^{d}$. Integrating an initial residual from $\mathbf{h}_{0}$ guarantees that the final representation of each token preserves at least a $\lambda$ portion of the information from the input layer. Here, $0<\lambda <1$ is a crucial factor when multiple layers are involved. Practically, we treat $\lambda$ as a hyperparameter, and setting it to a reasonable value, such as 0.2, ensures that the final token representation incorporates a substantial part of the input token feature, thereby reducing over-smoothness throughout the layers. 
We present a theoretical analysis in Appendix~\ref{app:proof}. In the following, we illustrate how our proposed framework, \model, can be applied to Adapter and LoRA, two most popular PEFT techniques. 

\stitle{Adapter-\model.}
Implementing the initial residual injection for adapters is straightforward. As illustrated in Fig.~\ref{fig:frame}(a), \model~adds the initial token representation $\mathbf{h}_{0}$ to a hidden state $\mathbf{h}$  at the entry point of the adapter (\ie, output from the preceding layer and input to the adapter), within each Transformer layer. 
This process is executed through a basic vector addition operation as follows.
\begin{align}
\mathbf{h}  \leftarrow& \tilde{\mathbf{h}} + f(\tilde{\mathbf{h}}\mathbf{W}_{\text{down}})\mathbf{W}_{\text{up}}\nonumber\\
&\text{s.t. } \tilde{\mathbf{h}} = (1-\lambda)\mathbf{h} + \lambda \mathbf{h}_{0},
    \label{eq:adapter-our}
\end{align}
where $0<\lambda<1$ is a hyper-parameter used to control the strength of the initial residual.

\stitle{LoRA-\model.}
In each LoRA module at every Transformer layer, the input to its update, $\Delta\mathbf{W}$, is solely the hidden state $\mathbf{h}$ from the preceding layer, with $\Delta\mathbf{W}$ being approximated by low-rank matrices. In LoRA-\model, we introduce a modification to the input to $\Delta\mathbf{W}$, which becomes a combination of $\mathbf{h}$ and $\mathbf{h}_{0}$, as follows.
\begin{align}
   \mathbf{h} \leftarrow & \mathbf{h} \mathbf{W} +  s \cdot \tilde{\mathbf{h}} \mathbf{W}_{\text{down}}\mathbf{W}_{\text{up}} \nonumber
   \\
   &\text{s.t. } \tilde{\mathbf{h}} = (1-\lambda)\mathbf{h} + \lambda \mathbf{h}_{0}.
\end{align}


\section{Experiments}

In this section, we perform extensive experiments across various benchmarks and language models.

\subsection{Datasets} 

Our study encompasses a thorough empirical examination of 22 benchmark datasets, categorized into three distinct problem areas as follows.

\stitle{Arithmetic reasoning.}  
(1) GSM8K \cite{cobbe2021training}: linguistically varied grade-school math word problems created by skilled problem writers.
(2) AQuA \cite{ling2017program}: algebraic word problems with natural language explanations.
(3) MAWPS \cite{koncel2016mawps}: a variety of arithmetic and algebra word problems of different complexities.
(4) SVAMP \cite{patel2021nlp}: arithmetic word problems aimed at students up to the 4th grade, derived by making minor modifications to an existing problem set.
PEFT techniques adopt a supervised fine-tuning (SFT) setting \cite{ouyang2022training}, 
where the supervision is derived from Math10K \cite{hu2023llm}, which comprises the training sets of GSM8K, AQuA, and MAWPS. The pre-trained model is fine-tuned on the examples in Math10K to replicate their styles and characteristics.

\stitle{Commonsense reasoning.}
(1) BoolQ \cite{clark2019boolq}: Yes/no questions, originating from natural, unrestricted environments.
(2) PIQA \cite{bisk2020piqa}: Questions requiring physical commonsense for resolving two possible solutions.
(3) SIQA \cite{sap2019social}:  Questions focusing on the understanding the social implications of human actions.
(4) HellaSwag: Commonsense natural language inference questions with various endings to complete a given context.
(5) WinoGrande \cite{sakaguchi2021winogrande}: A fill-in-the-blank task with binary choices, demanding commonsense reasoning to select the appropriate option.
(6) ARC-c and (7) ARC-e \cite{clark2018think}: The Challenge and Easy sets of the ARC dataset, featuring genuine grade-school level science questions in multiple-choice format.
(8) OBQA: Questions that necessitate multi-step reasoning, additional common and commonsense knowledge, and comprehensive text understanding.
To perform SFT, we employ a training set named Commonsense170K \cite{hu2023llm}, which is tailored for enhancing commonsense reasoning capabilities. It includes the training sets from the above eight commonsense reasoning datasets.

A summary of the datasets on arithmetic and commonsense reasoning is presented in Table~\ref{tab:data}.

\stitle{GLUE.} The General Language Understanding Evaluation Benchmark \cite{wang2018glue}
encompasses eight corpora for various natural language understanding tasks: CoLA, SST-2, MRPC, STS-B, QQP, MNLI, QNLI, and RTE.

\begin{table}[t]
  \centering
  \small
  \begin{tabular}{ccccc}
    \toprule
    Dataset & Domain & \# train & \# test & Answer \\
    \midrule
    GSM8K & Math & 8.8K & 1,319 & Number \\
    AQuA & Math & 100K & 254 & Option \\
    MAWPS & Math & 1.9k & 238 & Number \\
    SVAMP & Math & - & 1,000 & Number \\
    BoolQ & CS & 9.4K & 3,270 & Yes/No \\
    PIQA & CS & 16.1K & 1,838 & Option \\
    SIQA & CS & 33.4K & 1,954 & Option \\
    HellaSwag & CS & 39.9K & 10,042 & Option \\
    WinoGrande & CS & 63.2K & 1,267 & Option \\
    ARC-e & CS & 1.1K & 2,376 & Option \\
    ARC-c & CS & 2.3K & 1,172 & Option \\
    OBQA & CS & 5.0K & 500 & Option \\
    \bottomrule
  \end{tabular}
  \caption{Datasets on arithmetic reasoning (Math) or commonsense reasoning (CS).}
  \label{tab:data}
\end{table}





\begin{table*}[t]
\small
\centering
\addtolength{\tabcolsep}{-0pt}
\begin{tabular}{c|l|c|cccc|cc}
\hline
 PLM&PEFT method & \# Params.~tuned & GSM8K & AQuA & MAWPS & SVAMP  & Overall &Improv. \\ 
\hline
 GPT-3.5 (\text{175B})$^{*}$& $-$&$-$ &56.4&38.9&87.4&69.9&63.2&$-$ \\
\hline
\multirow{4}{*}{GPT-J (\text{6B})} 
& Adapter$^{*}$  &112M&14.3&20.5&62.2&38.1&33.8&$-$ \\
& Adapter-SIBO &112M&19.0&18.9&72.7&45.9&39.1&15.7\%  \\
& LoRA$^{*}$ &35M&17.4&21.3&70.2&41.0&37.5&$-$  \\
& LoRA-SIBO &35M&22.4&20.5&77.7&49.7&42.6&13.6\% \\
\hline
\multirow{4}{*}{LLaMA (\text{7B})} 
& Adapter$^{*}$  &128M&33.3&15.0&77.7&52.3&44.6&$-$  \\
& Adapter-SIBO &128M&33.1&18.9&80.3&48.0&45.1&1.1\%  \\
& LoRA$^{*}$ &40M&37.5&18.9&79.0&52.1&46.9&$-$  \\
& LoRA-SIBO &40M&37.8&18.5&82.8&50.7&47.5&1.3\%  \\
\hline
\multirow{4}{*}{LLaMA (13B)} 
& Adapter$^{*}$  &200M&44.0&22.0&78.6&50.8&48.9&$-$  \\
& Adapter-SIBO &200M&43.2&22.4&82.4&52.9&50.2&2.7\%  \\
& LoRA$^{*}$ &62.5M&47.5&18.5&83.6&54.6&51.1&$-$  \\
& LoRA-SIBO &62.5M&47.0&20.5&84.0&57.6&52.3&2.3\%  \\
\hline
\end{tabular}
\caption{Performance of LLMs with different PEFT methods on arithmetic reasoning, using GPT-3.5 with zero-shot CoT as a reference point.
$^{*}$ indicates results from prior work by \citet{hu2023llm}, where the exact same experimental setup and evaluation protocols are adopted. Improvement is calculated relative to the counterpart without \model. 
}
\label{tab:math}
\end{table*}

\begin{table*}[t]
 \small
\centering
\addtolength{\tabcolsep}{-.5pt}
\begin{tabular}{l|cccccccc|cc}
\hline
 Method & BoolQ	&PIQA	&SIQA	&HellaSwag	&WinoGrande	&ARC-e	&ARC-c	&OBQA  & Overall& Improv. \\ 
\hline

Adapter$^{*}$  &62.1&63.5&72.3&30.6&68.0&63.9&48.1&63.8&59.0&$-$  \\
Adapter-SIBO &62.2&73.0&73.0&48.7&67.8&65.5&51.9&65.6&63.5&7.6\%  \\
LoRA$^{*}$ &62.4&68.6&49.5&43.1&57.3&43.4&31.0&46.6&50.2&$-$  \\
LoRA-SIBO &63.9&70.3&71.0&47.8&67.2&63.3&48.7&63.8&62.0&23.5\%  \\
\hline
\end{tabular}
\caption{Performance of GPT-J (\text{6B}) with different PEFT methods on commonsense reasoning. 
$^{*}$ indicates results from prior work \cite{hu2023llm}, where the exact same experimental setup and evaluation protocols are adopted. 
}
\label{tab:common sense}
\end{table*}

\begin{table*}[t]
\small
\centering
\addtolength{\tabcolsep}{-3pt}
\begin{tabular}{l|c|cccccccc|cc}
\hline
 Method& \# Params.~tuned  & CoLA & SST-2 & MRPC & STS-B & QQP & MNLI & QNLI & RTE & Overall \\ 
\hline
FT$^{*}$& 340.0M & 62.8\phantom{$_{\pm 0.0}$} & 94.1\phantom{$_{\pm 0.0}$} & 91.9\phantom{$_{\pm 0.0}$} & 89.8\phantom{$_{\pm 0.0}$} & 87.6\phantom{$_{\pm 0.0}$} & 86.5\phantom{$_{\pm 0.0}$}  & 93.5\phantom{$_{\pm 0.0}$} & 71.8\phantom{$_{\pm 0.0}$} & 84.8  \\\hline
Adapter  & 6.0M &62.1$_{\pm 1.1}$&93.7$_{\pm 0.2}$&90.4$_{\pm 0.5}$&90.2$_{\pm 0.2}$&88.3$_{\pm 0.3}$&85.9$_{\pm 0.1}$&92.2$_{\pm 0.2}$&71.5$_{\pm 2.4}$&84.3 \\
Adapter-SIBO &6.0M&63.1$_{\pm 1.2}$&94.6$_{\pm 0.2}$&90.9$_{\pm 0.1}$&90.2$_{\pm 0.1}$&88.3$_{\pm 0.2}$&86.0$_{\pm 0.1}$&92.4$_{\pm 0.3}$&73.2$_{\pm 1.1}$ & 84.8 \\
LoRA &0.8M&60.1$_{\pm 1.0}$&93.6$_{\pm 0.3}$&90.3$_{\pm 0.3}$&89.6$_{\pm 0.1}$&87.8$_{\pm 0.1}$&85.5$_{\pm 0.1}$&92.1$_{\pm 0.3}$&71.1$_{\pm 0.8}$& 83.8 \\

LoRA-SIBO &0.8M&61.6$_{\pm 0.8}$&93.8$_{\pm 0.2}$&90.8$_{\pm 0.1}$&89.9$_{\pm 0.1}$&87.7$_{\pm 0.2}$&85.6$_{\pm 0.2}$&92.2$_{\pm 0.2}$&71.8$_{\pm 1.8}$ & 84.2\\

\hline
\end{tabular}
\caption{Performance of BERT-large with different PEFT methods on the
GLUE benchmark. 
$^{*}$ indicates results from prior work \cite{zaken2022bitfit}, where the exact same experimental setup and evaluation protocols are adopted. We report mean (and standard deviation) of the performance over three different runs. 
}
\label{tab:glue}
\end{table*}

\subsection{Implementations}

\stitle{Arithmetic and commonsense reasoning.} We use LLaMA (7B, 13B) \cite{touvron2023llama} and GPT-J (6B) \cite{wang2021gpt} as the foundational models, which are designed for natural language generation tasks.
We choose Adapter \cite{houlsby2019parameter} and LoRA \cite{hu2021lora} as baselines, and follow previous work \cite{hu2023llm} for the experimental setup and hyperparameters. 
In particular, for Adapter, we integrate it into the feed-forward layers with a bottleneck size of 256; 
for LoRA, we incorporate it into both the multi-head attention and feed-forward layers with rank 32.
For Adapter-\model\ and LoRA-\model, we inject the initial residual into the modules at the feed-forward layers only, and choose $\lambda\in \{0.1,0.2,0.3\}$ empirically while retaining other settings in the vanilla Adapter and LoRA.  
More details on the experimental setup can be found in Appendix~\ref{app:hyper parameter}.

\stitle{GLUE.} We use BERT-large \cite{devlin2018bert} as the backbone. While larger models have recently surpassed BERT on the GLUE benchmark, BERT continues to be favored for its efficiency. Moreover, it is relatively easy to perform full-model fine-tuning (FT) on BERT, enabling a direct comparison between FT and PEFT techniques.
For Adapter, we apply the typical setting \cite{houlsby2019parameter} where adapter layers are added after the multi-head attention and feed-forward layers; for LoRA, we follow previous work \cite{hu2021lora} and apply to weights $\mathbf{W}_{q}$ and $\mathbf{W}_{v}$ with rank 8.
For Adapter-\model, we inject the initial residual to the adapter modules after the self-attention layers;
for LoRA-\model, we inject the initial residual to all LoRA modules. For both \model\ approaches, we choose $\lambda\in \{0.1,0.2,\ldots,0.7\}$ empirically while following previous work \cite{houlsby2019parameter, hu2021lora} to set other hyperparameters.

\subsection{Performance comparison}

We evaluate the performance of \model\ in comparison to baselines across the three problem areas. 



\stitle{Arithmetic reasoning.} We compare the performance of Adapter and LoRA with or without SIBO, by performing PEFT on the pre-trained LLaMA and GPT-J models using the Math10K dataset. We then test the fine-tuned models  across the test set of the four math reasoning datasets. As a standard reference \cite{hu2023llm}, we further compare to the GPT-3.5 model (text-Davinci-003 version), which employs zero-shot Chain of Thought (CoT) \cite{kojima2022large}.

As reported in Table~\ref{tab:math}, 
the 175B-parameter GPT-3.5 model demonstrates superior accuracy over other LLMs. 
Despite this, LoRA-\model\ applied on LLaMA (13B) has reached a performance level comparable to that of GPT-3.5 with only a small gap. Compared to the counterparts without \model, SIBO has achieved notable improvements: 2.3\%--2.7\% on LLaMA (13B), and 1.1\%--1.3\% on LLaMA (7B). The smaller improvements on the 7B model can be attributed to the less pronounced over-smoothing issue in smaller models with fewer layers, indirectly underscoring the necessity to address over-smoothing in deeper models. Meanwhile, \model\ achieves up to 15.7\% improvement on the weaker GPT-J, significantly reducing the gap from LLaMA (7B).


Moreover, we observe enhancements by \model\ in both in- and out-of-distribution scenarios. The dataset utilized for fine-tuning, Math10K, encompasses the training sets from GSM8K, AQuA, and MAWPS, excluding SVAMP. It can be observed that \model\ not only enhances the performance of PEFT methods on the first three datasets in an in-distribution setting, but also extends the improvements to SVAMP, an out-of-distribution scenario, demonstrating the robustness and generalizability of our methodology.


\stitle{Commonsense reasoning.}
Next, we investigate the performance of \model\ for commonsense reasoning tasks. 
Table~\ref{tab:common sense} presents a comparative analysis of the PEFT methods applied to GPT-J (\text{6B}). 
It is evident that \model\ consistently and significantly enhances the performance of Adapter and LoRA across eight diverse corpora/tasks, with average improvements ranging between 7.6\% and 23.5\%.

\stitle{GLUE.}
Lastly, we present the results on the GLUE benchmark in Table~\ref{tab:glue}, using BERT-large as the backbone model.  \model\ consistently outperforms the vanilla PEFT methods across eight diverse datasets/tasks. Notably, the effectiveness of the Adapter-\model\ even matches that of full-model fine-tuning (FT).
More experimental results using an alternative pre-trained model, RoBERTa-large, are in Appendix~\ref{app:roberta glue}.

\subsection{Analyses}
In this section, we first analyze the optimal placement of the initial residual for Adapter and LoRA. Following that, we examine the effect of the sole hyperparameter $\lambda$ we introduced. Then, we explore the overhead incurred  by \model. Finally, we visualize the role of the initial residual in mitigating the over-smoothing issue, and present a case study. In these studies, we employ BERT-large as the backbone on the CoLA and STS-B datasets.

\stitle{Placement.}
In this section, we investigate the placement of initial residual for PEFT modules.

For Adapter, each Transformer layer employs two adapter modules, positioned respectively after the attention layer (ATT) and the feed-forward layer (FFN). The question arises: which position is more suitable for the injection of the initial residual? As shown in Table~\ref{tab:place adapter}, injecting the initial residual solely at the ATT position achieves almost identical performance to that at the FFN position. However, injecting initial residuals at both ATT and FFN results in a slight decline in performance. This suggests that injecting the initial residual once per Transformer layer is sufficient, as excessive injections can introduce noises.

For LoRA, each module involves two types of parameters: frozen pre-trained weights and learnable low-rank matrices. We explore whether it is necessary to inject the initial residual into both or solely into the low-rank matrices. Table~\ref{tab:place LoRA} reveals that injecting the initial residual only into the learnable low-rank matrices yields better results. A potential reason is that frozen weights do not integrate well with the layer's hidden state and the initial residual. 

\begin{table}[t]
    \small
 \centering 
\begin{tabular}{c|cc|c}  
		\hline
		 Placement&CoLA&STS-B&Average \\\hline
		ATT &63.1$_{\pm 1.2}$&90.2$_{\pm 0.2}$&76.7 \\
            FFN &63.3$_{\pm 0.6}$&90.1$_{\pm 0.1}$&76.7 \\
            Both &61.8$_{\pm 1.5}$&90.0$_{\pm 0.1}$&75.9 \\
		\hline
\end{tabular}
\caption{Initial residual placement for Adapter.} 
\label{tab:place adapter}
\end{table}

\begin{table}[t]
    \small
 \centering 
\begin{tabular}{c|c|c|c}  
		\hline`
		 Placement&CoLA&STS-B&Average \\\hline
		Low-rank matrices  &61.6$_{\pm 0.8}$&89.9$_{\pm 0.1}$&75.8\\
            + Pre-trained weights  &61.0$_{\pm 0.8}$&89.7$_{\pm 0.3}$&75.1\\
		\hline
\end{tabular}
\caption{Initial residual placement for LoRA.} 
\label{tab:place LoRA}
\end{table}

\begin{figure}[t]
   \subfigure[CoLA]{
   \centering
   \includegraphics[width=0.47\linewidth]{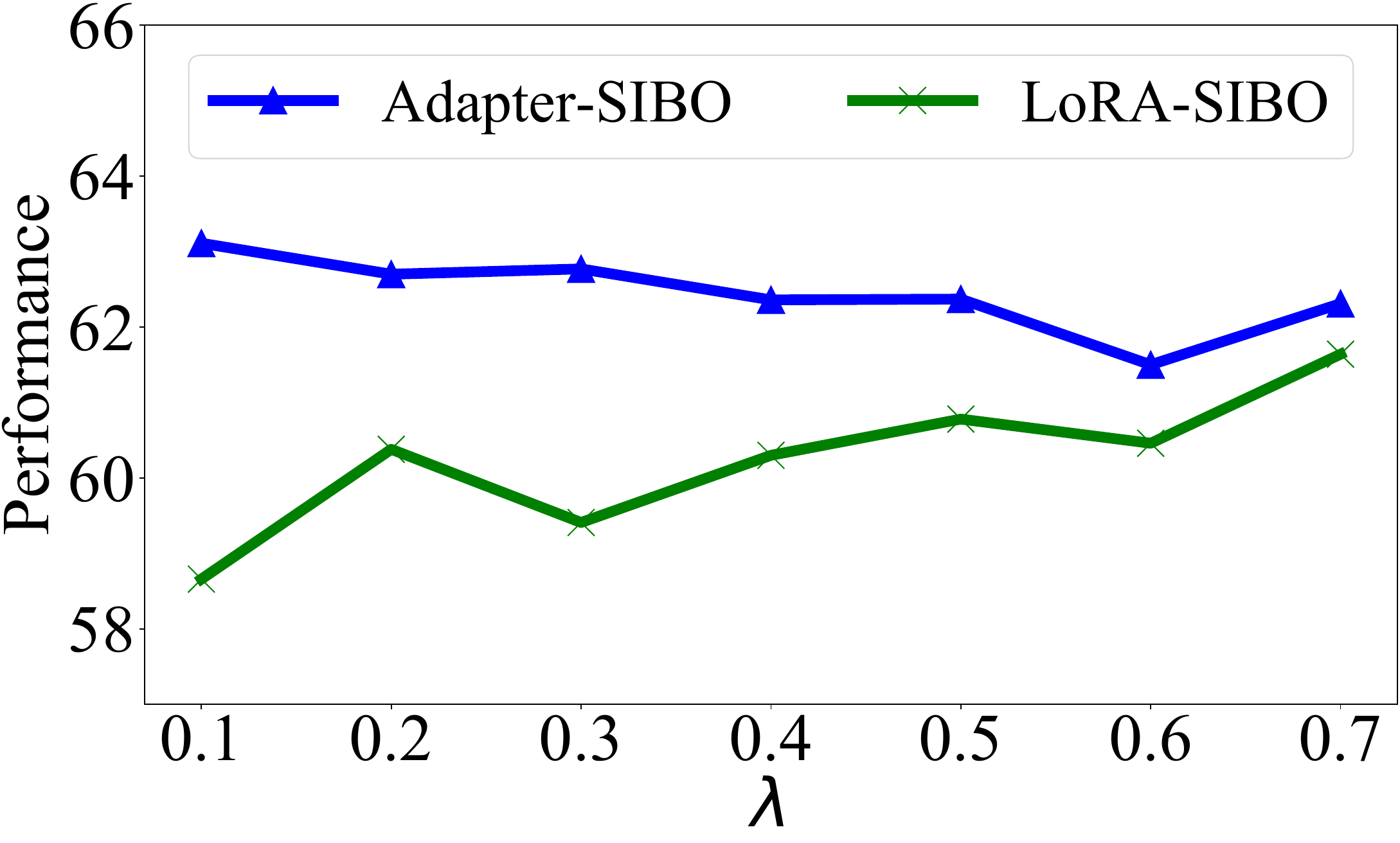}
   }
   \subfigure[STS-B]{
   \centering
   \includegraphics[width=0.47\linewidth]{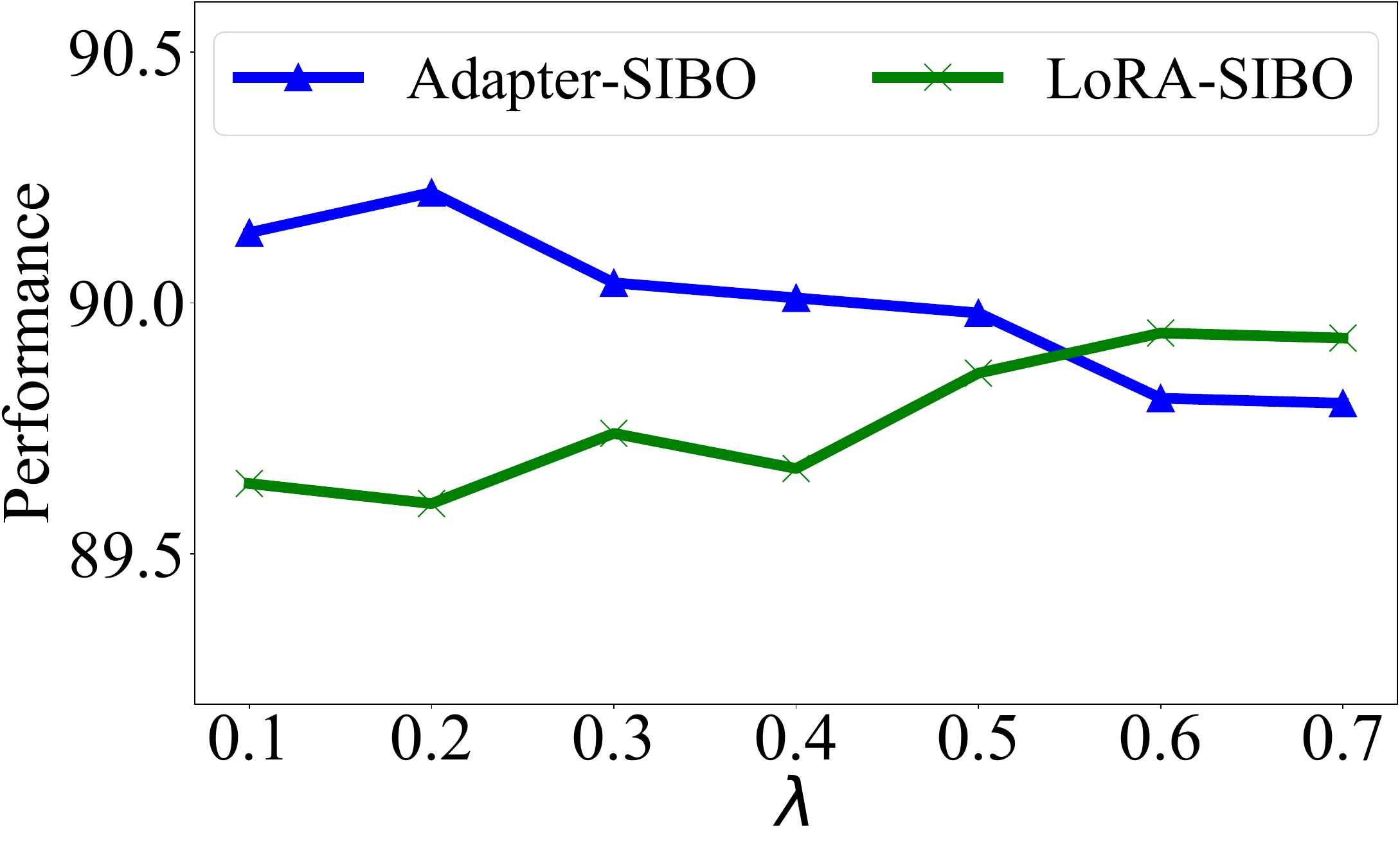}
   }
    \vspace{-2mm}
	\caption{Impact of initial residual portion $\lambda$.}
	\label{fig:lambda}
\end{figure}

\stitle{Impact of $\lambda$.}
\model\ only introduces one new hyper-parameter, $\lambda$, which balances the trade-off between the hidden state and the initial residual. The selection of $\lambda$ also guarantees the minimum portion of input features preserved in the final token representation, directly mitigating the over-smoothing issue. Hence, we investigate the optimal value for $\lambda$, varying it between 0.1 and 0.7. As illustrated in Fig.~\ref{fig:lambda}, for Adapter, a lower $\lambda$ value, such as 0.2, is generally more effective. While it is crucial to ensure that the final representation of each token maintains a minimum portion of $\lambda$ from the input layer across multiple stacked layers, this proportion should not be excessively large to avoid compromising the learning capacity of Adapter. For LoRA, the hidden state is fed to both the pre-trained weights and the low-rank matrices, implying that the effect of $\lambda$ ratio is naturally ``halved''. In other words, a $\lambda$ value of 0.6 for LoRA is roughly equivalent to a $\lambda$ value of 0.3 for Adapter. Therefore, the optimal value of $\lambda$ in LoRA is larger than in Adapter, occurring around 0.6--0.7.

Despite the effort to select $\lambda$, our approach remains pragmatic and resource-conscious. For smaller models like BERT, extensive tuning over a large range of values for $\lambda$ is feasible.
In the context of larger models, such as LLaMA (13B), a smaller range of values $\lambda\in\{0.1,0.2,0.3\}$ have been considered in our tuning, which still yields significant improvements across various tasks, 
as shown in Tables~\ref{tab:llama13b adapter lambda} and \ref{tab:llama13b lora lambda}.
Notably, many non-optimal values of $\lambda$ could still result in significant performance gains, underscoring the efficacy and robustness of \model\ without extensive hyperparameter tuning. 

\begin{table}[t]
    \small
 \centering
 \addtolength{\tabcolsep}{-3pt}
    \begin{tabular}{c|ccccc}
        \hline
        Methods & GSM8k & AQuA & MAWPS & SVAMP & Avg. \\ \hline
        Adapter & \textbf{44.0} & 22.0 & 78.6 & 50.8 & 48.9 \\ \hline
        SIBO ($\lambda$ = 0.1) & 42.7 & 17.3 & \textbf{83.6} & \textbf{55.9} & 49.9  \\ 
        SIBO ($\lambda$ = 0.2) & 42.3 & 20.1 & 81.9 & 55.3 & 49.9 \\ 
        SIBO ($\lambda$ = 0.3) & 43.2 & \textbf{22.4} & 82.4 & 52.9 & \textbf{50.2}  \\ \hline
    \end{tabular}
    \caption{Impact of $\lambda$, with LLaMA (13B) as the backbone, and adapter as the PEFT method.}
    \label{tab:llama13b adapter lambda}
\end{table}

\begin{table}[t]
    \small
 \centering
 \addtolength{\tabcolsep}{-3pt}
    \begin{tabular}{c|ccccc}
        \hline
        Methods & GSM8k & AQuA & MAWPS & SVAMP & Avg. \\ \hline
        LoRA & 47.5 & 18.5 & 83.6 & 54.6 & 51.1\\ \hline
        SIBO ($\lambda$ = 0.1) & 47.0 & 20.5 & \textbf{84.0} & 57.6 & \textbf{52.3}  \\ 
        SIBO ($\lambda$ = 0.2) & 46.6 & 19.3 & \textbf{84.0} & \textbf{57.8} & 51.9  \\
        SIBO ($\lambda$ = 0.3) & \textbf{47.8} & \textbf{21.3} & 83.2 & 52.9 & 51.3  \\ \hline
    \end{tabular}
    \caption{Impact of $\lambda$, with LLaMA (13B) as the backbone, and LoRA as the PEFT method.}
    \label{tab:llama13b lora lambda}
\end{table}

\stitle{Complexity.}
\model\ is remarkably efficient, involving only an additional summation operation with the initial residual vector at each Transformer layer, without introducing any extra parameter. To demonstrate its efficiency, we compare the number of floating point operations (FLOPs) and the wall-clock time for fine-tuning and testing. As shown in Table~\ref{tab:complexity}, the overhead of summing the initial residual vector only marginally increases the FLOPs. 
Moreover, the wall-clock time is almost identical to that of the vanilla PEFT methods, which does not include the initial residual, highlighting the simplicity and efficiency of \model. 

\begin{table}[t]
\small
 \centering
\addtolength{\tabcolsep}{-2pt}
\begin{tabular}{@{}l|crrr@{}}
\hline
\multirow{2}{*}{Methods} & \# &\multirow{2}{*}{FLOPs}& \multicolumn{1}{c}{CoLA} & \multicolumn{1}{c}{STS-B}  \\
& Params.&& Time (s) & Time (s)  \\
\hline
Adapter       &6.0M&6,291,456&108.3$_{\pm 0.9}$&82.7$_{\pm 0.6}$ \\
Adapter-SIBO        &6.0M&6,389,760&110.0$_{\pm 1.0}$&80.7$_{\pm 1.5}$  \\
LoRA &0.8M&835,584&86.3$_{\pm 0.6}$& 55.7$_{\pm 0.5}$     \\
LoRA-SIBO &0.8M&884,736&90.0$_{\pm 1.0}$ & 52.7$_{\pm 7.5}$ \\
\hline
\end{tabular}
\caption{Complexity analysis. Time includes fine-tuning one epoch and then testing, averaged over three runs.
}
\label{tab:complexity}
\end{table}



\begin{figure}[t]
   \subfigure[Adapter]{
   \centering
   \includegraphics[width=0.48\linewidth]{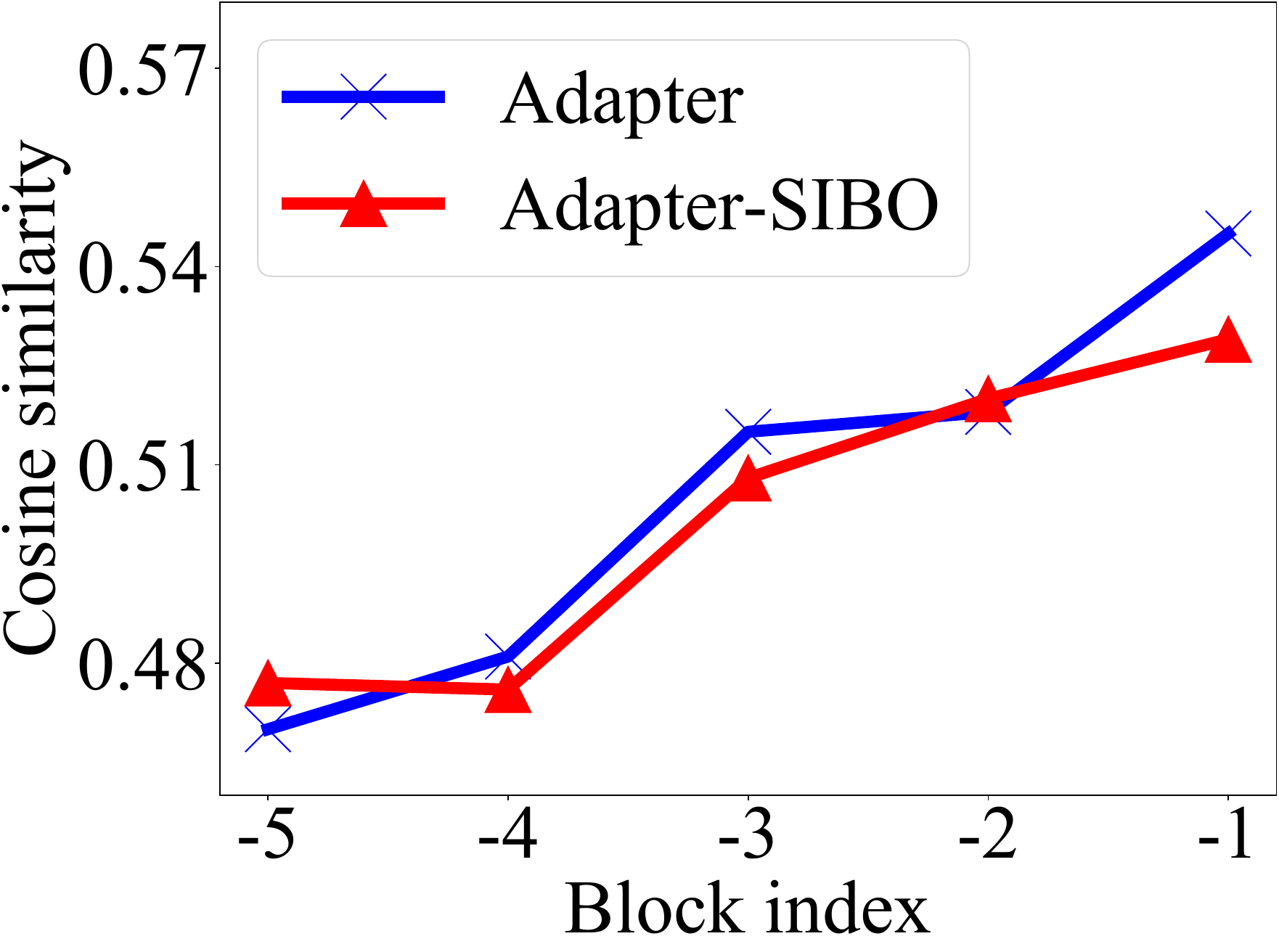}
   }%
   \subfigure[LoRA]{
   \centering
   \includegraphics[width=0.48\linewidth]{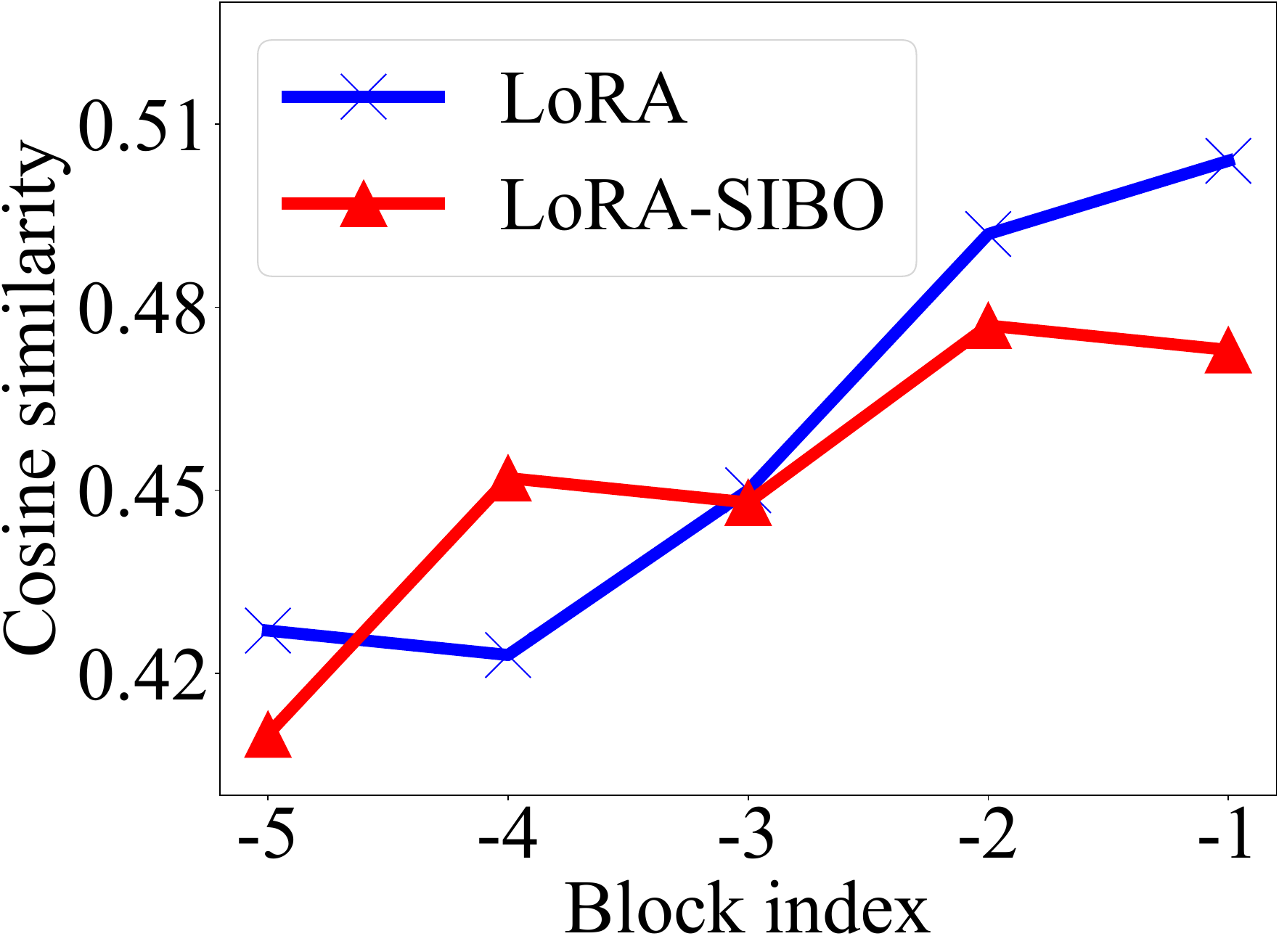}}
   \vspace{-2mm}
	\caption{Token-wise similarity in last five layers computed
from PEFT methods, with and without \model.}
	\label{fig:tokensim curve}
\end{figure}

\begin{figure}[t]
   \subfigure[LoRA]{
   \centering
   \includegraphics[width=0.48\linewidth]{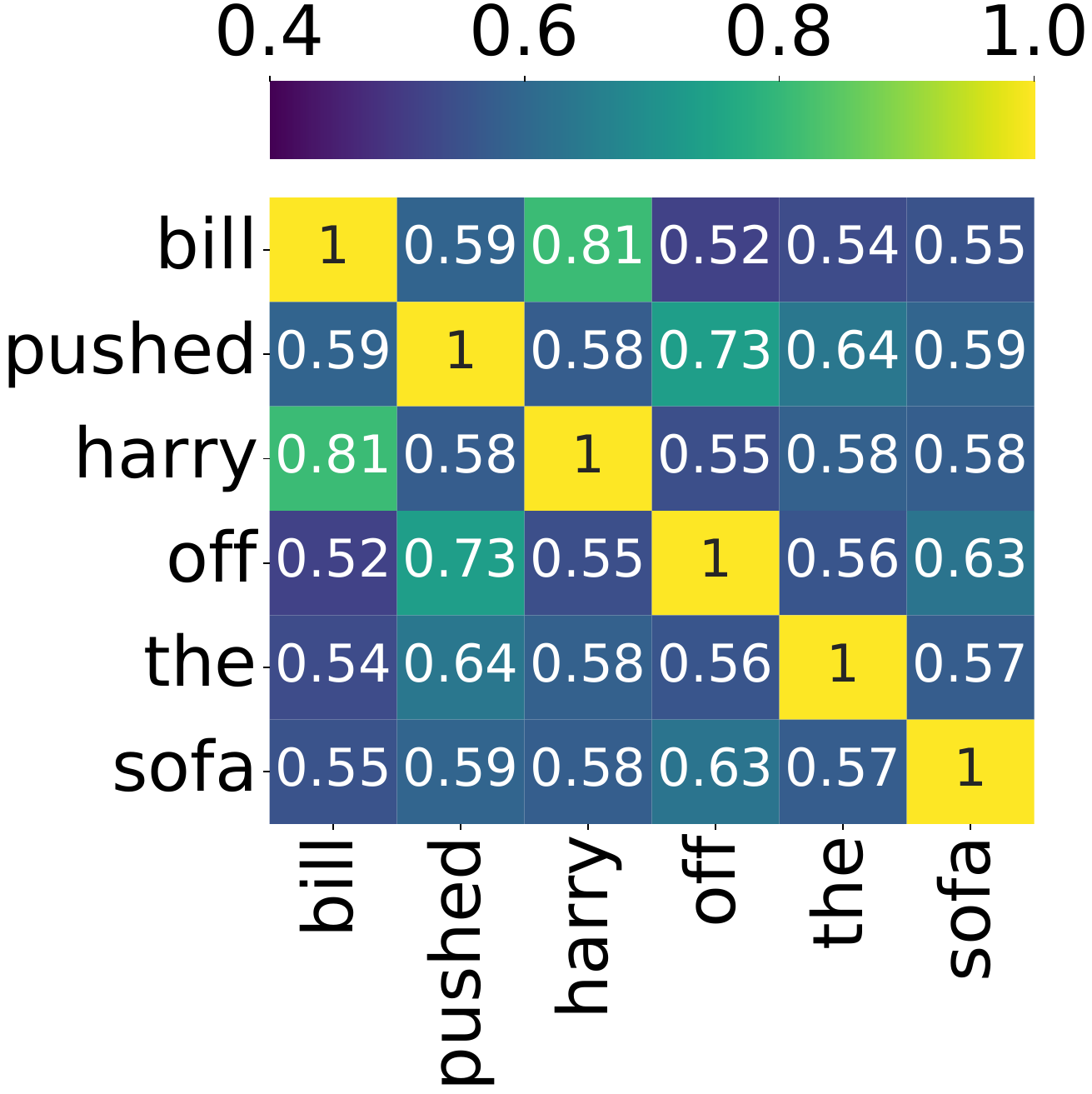}
   }%
   \subfigure[LoRA-SIBO]{
   \centering
   \includegraphics[width=0.48\linewidth]{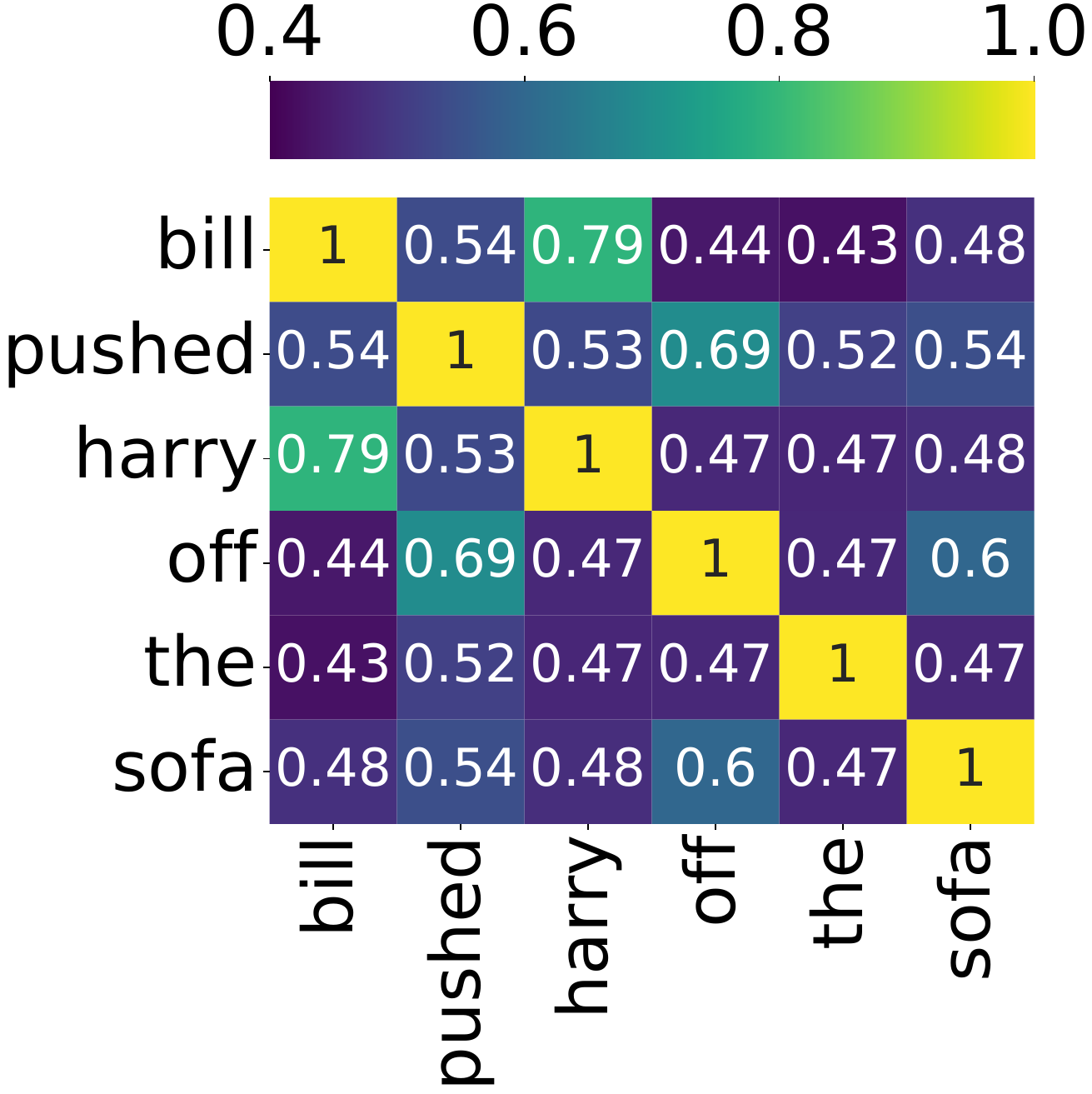}}
   \vspace{-2mm}
	\caption{Heatmap of token-wise similarity in the last layer computed from LoRA and LoRA-\model, on a sentence randomly sampled from the test set of CoLA. } 
	\label{fig:tokensim heat}
\end{figure}

\stitle{Visualizations of over-smoothing.}
The thesis of the work is to employ the initial residual to alleviate over-smoothing. To examine whether \model\ effectively reduces over-smoothing, we conduct experiments comparing the token-wise cosine similarity, as defined in Eq.~\ref{eq:tokenwise_sim}, in the last five layers of the language model after applying PEFT methods with or without \model. As observed in Figs.~\ref{fig:tokensim curve} and \ref{fig:tokensim heat}, the token-wise similarity generally decreases when \model\ is applied alongside  Adapter and LoRA. In essence, \model\ has lessened the degree of over-smoothing, leading to better task performance.

\stitle{Qualitative case study.}
Finally, we supplement our quantitative findings with qualitative analysis in a case study.
Table~\ref{tab:quality} presents a question sampled from SVAMP, showcasing responses from ChatGPT 3.5, as well as LLaMA (13B) using various PEFT methods. While ChatGPT is generally robust, it is not infallible and has provided an incorrect answer in this instance. Similarly, the answers generated by Adapter and LoRA were found to be erroneous, mainly due to their initial confusion between the related concepts of \texttt{cracker} and \texttt{snack}.

In contrast, LoRA-\model\ shows an improvement by correctly distinguishing between the two concepts at least in the initial phase of reasoning. Adapter-\model\ outperforms all by maintaining clarity on the two concepts throughout, leading to the correct answer. This improvement underscores the benefit of addressing over-smoothness, which enhances word discrimination and thus provides a clearer distinction between closely related concepts such as \texttt{cracker} and \texttt{snack}, improving overall performance. Moreover, the responses from Adapter-\model\ and LoRA-\model\ appear more detailed and of higher quality than those from the vanilla Adapter and LoRA. 
Further qualitative examples are presented in Appendix~\ref{app:qualitative}.

\begin{table*}[t]
\small
\centering 
\resizebox*{\textwidth}{!}{%
\renewcommand{\arraystretch}{0.7}
\begin{tabular}{@{}p{2.1cm}|p{14cm}@{}}
		\toprule
		 \multirow{2}{*}{\textbf{Question}}  & \texttt{Matthew gave equal numbers of crackers and cakes to his 4 friends. If he had 32 crackers and 98 cakes initially. How many crackers did each person eat?}  \\\midrule
		\multirow{9}{*}{\textbf{ChatGPT$^{*}$}}  & \texttt{To calculate the number of crackers each person ate, we need to find the number of crackers that each friend received initially and subtract it from the total number of crackers Matthew had.} \\
& \texttt{Here are the steps:} \\
& \texttt{1. Divide the total number of crackers by the number of friends (4) to find the number of crackers each friend received initially: 32 / 4 = 8 crackers per friend.} \\
& \texttt{2. Since we want to find the number of crackers each person ate, we need to subtract the initial number of crackers per friend from the total number of crackers: 32 - 8 = 24 crackers.} \\\midrule
            \multirow{3}{*}{\textbf{LLaMA (\text{13B})}}  & \texttt{1. Divide the total number of crackers by the number of friends: 32 / 4 = 8} \\
& \texttt{2. Divide the total number of cakes by the number of friends: 98 / 4 = 24} \\
            \multirow{1}{*}{\textbf{+ Adapter$^{*}$}}   & \texttt{3. Add the number of crackers and cakes each friend received: 8 + 24 = 32} \\
& \texttt{4. Therefore, each person ate 32 crackers and 24 cakes.} \\
&\texttt{Answer: Each person ate 32 crackers.}\\\midrule
            \multirow{5}{*}{\textbf{LLaMA (\text{13B})}}   & \texttt{1. Find the total number of crackers and cakes: 32 + 98 = 130} \\
            & \texttt{2. Divide the total number of crackers by the number of friends: 32 $\div$ 4 = 8}\\
            &\texttt{3. Divide the total number of cakes by the number of friends: 98 $\div$ 4 = 22}\\
            \multirow{1}{*}{\textbf{+ Adapter-\model}}  
            &\texttt{4. Since Matthew gave equal numbers of crackers and cakes, each friend ate the same number of crackers and cakes.} \\
            &\texttt{5. Therefore, each person ate 8 crackers and 22 cakes.}\\
            &\textbf{\texttt{Therefore, each person ate \underline{8} crackers.}}\\
            \midrule
            \multirow{3}{*}{\textbf{LLaMA (\text{13B})}}  & \texttt{1. Divide the total number of crackers by the number of friends: 32 / 4 = 8} \\
& \texttt{2. Divide the total number of cakes by the number of friends: 98 / 4 = 24} \\
            \multirow{1}{*}{\textbf{+ LoRA$^{*}$}} & \texttt{3. Add the number of crackers and cakes each person received: 8 + 24 = 32} \\\midrule
            \multirow{6}{*}{\textbf{LLaMA (\text{13B})}}  &\texttt{1. Find the total number of friends Matthew gave crackers and cakes to: 4 friends} \\
            &\texttt{2. Find the total number of crackers and cakes Matthew gave away: 32 crackers + 98 cakes = 130 snacks} \\
            \multirow{2}{*}{\textbf{+ LoRA-\model}}  
            &\texttt{3. Divide the total number of snacks by the number of friends: 130 snacks $\div$ 4 friends = 32.5 snacks per person} \\
            &\texttt{4. Round down to the nearest whole number since Matthew can't give away a fraction of a cracker: 32 crackers per person} \\
            &\texttt{Therefore, each person ate 32 crackers.}\\
		\hline
\end{tabular}}
\caption{An example question from SVAMP, with the ground-truth answer \underline{\texttt{\textbf{8}}}. We show the output of ChatGPT and various PEFT methods on LLaMA (13B). $^{*}$ indicates results from prior work \cite{hu2023llm}.} 
\label{tab:quality}
\end{table*}

\section{Related work}

We focus on the discussion of parameter-efficient fine-tuning (PEFT) of pre-trained language models.
There exists three main categories of methods, including prompt-based learning, adapters, and reparametrization methods. However, none of the existing approaches is designed to alleviate the over-smoothing issue.

Prompt-based learning extends the identification of the ideal discrete (hard) prompt into the optimization of a continuous (soft) prompt instead. \citet{lester2021power} have introduced the idea of prompt tuning, which involves attaching a trainable tensor as a prefix to the input embeddings. Similarly, \citet{li2021prefix} have developed an independent method known as prefix tuning, which integrates soft prompts into the hidden states across all layers. Another technique, Intrinsic Prompt Tuning \cite{qin2021exploring}, utilizes an autoencoder to both compress and decompress the soft prompt at the cost of limiting the sequence length.

Adapters exist in parallel and serial forms. Parallel adapters \cite{he2021towards} integrate additional learnable modules alongside various layers of the core model. A different strategy, termed Ladder Side-Tuning \cite{sung2022lst}, focuses on developing a streamlined auxiliary network akin to a ladder. This auxiliary network receives intermediate activations from the main network via direct shortcut pathways, referred to as ladders. In contrast, serial adapters insert these modules sequentially between specific layers. \citet{houlsby2019parameter} add fully connected networks after both the attention and feed-forward layers in the Transformer model. \citet{pfeiffer2020mad} have demonstrated that inserting an adapter only after the self-attention layer can yield performance comparable to using two adapters per transformer block, whereas AdaMix \citet{wang2022adamix} employs multiple serial adapters in a mixture-of-experts approach. To further reduce computational complexity while preserving performance, Compacter \cite{henderson2021compacter} leverages the Kronecker product, low-rank matrices, and parameter sharing across layers for adapter weight generation.

Finally, reparametrization-based methods are designed to modify network weights through a low-rank approximation. This technique effectively minimizes the number of trainable parameters without compromising the representational capacity of high-dimensional matrices. The work on Intrinsic SAID \cite{aghajanyan2021intrinsic} examines the essential dimensionality of fine-tuning within a low-rank framework. On the other hand, LoRA \cite{hu2021lora} models its update to a pre-trained weight matrix through a low-rank decomposition.
Building on this, \citet{edalati2022krona} enhance the matrix decomposition feature of LoRA by incorporating the Kronecker product into their method. 


\section{Conclusion}
We present a novel framework \model, a \underline{SI}mple \underline{BO}oster to enhance parameter-efficient fine-tuning (PEFT) techniques for large pre-trained language models. Our core idea revolves around mitigating the over-smoothing issue, which is achieved by injecting an \emph{initial residual} into various PEFT modules at specific positions within the pre-trained models.
\model\ is straightforward and readily extensible to various state-of-the-art PEFT methods including Adapter and LoRA. Extensive experiments on 22 benchmark datasets across three problem areas demonstrate that \model\ effectively mitigate over-smoothing and significantly improves the performance of existing PEFT techniques.


\section{Limitations}
Our method is straightforward and effective, yet it has one limitation: selecting the optimal value for the hyperparameter $\lambda$ requires time and computational resources. This cost is manageable given that we only introduced one new hyperparameter, especially for medium-sized models. However, it may become prohibitive for very large models. A viable solution is to transform this hyperparameter into a continuous learnable parameter, allowing the model to autonomously determine the optimal weight for the initial residual.

\section*{Acknowledgements}
This research is supported by the Agency for Science, Technology and Research (A*STAR) under its AME Programmatic Funds (Grant No. A20H6b0151).
The authors wish to thank Dr. Lei Wang from Singapore Management University and Mr. Zhiqiang Hu from the Singapore University of Technology and Design for their valuable support of this work.



\bibliography{references}

\begin{thebibliography}{39}
\expandafter\ifx\csname natexlab\endcsname\relax\def\natexlab#1{#1}\fi

\bibitem[{Aghajanyan et~al.(2021)Aghajanyan, Gupta, and Zettlemoyer}]{aghajanyan2021intrinsic}
Armen Aghajanyan, Sonal Gupta, and Luke Zettlemoyer. 2021.
\newblock Intrinsic dimensionality explains the effectiveness of language model fine-tuning.
\newblock In \emph{Annual Meeting of the Association for Computational Linguistics and International Joint Conference on Natural Language Processing (ACL-IJCNLP)}, pages 7319--7328.

\bibitem[{Bisk et~al.(2020)Bisk, Zellers, Gao, Choi et~al.}]{bisk2020piqa}
Yonatan Bisk, Rowan Zellers, Jianfeng Gao, Yejin Choi, et~al. 2020.
\newblock Piqa: Reasoning about physical commonsense in natural language.
\newblock In \emph{AAAI Conference on Artificial Intelligence}, pages 7432--7439.

\bibitem[{Brunner et~al.(2019)Brunner, Liu, Pascual, Richter, Ciaramita, and Wattenhofer}]{brunner2019identifiability}
Gino Brunner, Yang Liu, Damian Pascual, Oliver Richter, Massimiliano Ciaramita, and Roger Wattenhofer. 2019.
\newblock On identifiability in transformers.
\newblock In \emph{International Conference on Learning Representations (ICLR)}.

\bibitem[{Chen et~al.(2023)Chen, Shou, Gong, Pei, Cao, Chang, Jiang, and Li}]{chen2023alleviating}
Nuo Chen, Linjun Shou, Ming Gong, Jian Pei, Bowen Cao, Jianhui Chang, Daxin Jiang, and Jia Li. 2023.
\newblock Alleviating over-smoothing for unsupervised sentence representation.
\newblock In \emph{Annual Meeting of the Association for Computational Linguistics (ACL)}, pages 3552--3566.

\bibitem[{Clark et~al.(2019)Clark, Lee, Chang, Kwiatkowski, Collins, and Toutanova}]{clark2019boolq}
Christopher Clark, Kenton Lee, Ming-Wei Chang, Tom Kwiatkowski, Michael Collins, and Kristina Toutanova. 2019.
\newblock Boolq: Exploring the surprising difficulty of natural yes/no questions.
\newblock In \emph{Conference of the North American Chapter of the Association for Computational Linguistics: Human Language Technologies (NAACL-HLT)}, pages 2924--2936.

\bibitem[{Clark et~al.(2018)Clark, Cowhey, Etzioni, Khot, Sabharwal, Schoenick, and Tafjord}]{clark2018think}
Peter Clark, Isaac Cowhey, Oren Etzioni, Tushar Khot, Ashish Sabharwal, Carissa Schoenick, and Oyvind Tafjord. 2018.
\newblock Think you have solved question answering? {Try ARC, the AI2 Reasoning Challenge}.
\newblock \emph{arXiv preprint arXiv:1803.05457}.

\bibitem[{Cobbe et~al.(2021)Cobbe, Kosaraju, Bavarian, Chen, Jun, Kaiser, Plappert, Tworek, Hilton, Nakano et~al.}]{cobbe2021training}
Karl Cobbe, Vineet Kosaraju, Mohammad Bavarian, Mark Chen, Heewoo Jun, Lukasz Kaiser, Matthias Plappert, Jerry Tworek, Jacob Hilton, Reiichiro Nakano, et~al. 2021.
\newblock Training verifiers to solve math word problems.
\newblock \emph{arXiv preprint arXiv:2110.14168}.

\bibitem[{Devlin et~al.(2019)Devlin, Chang, Lee, and Toutanova}]{devlin2018bert}
Jacob Devlin, Ming{-}Wei Chang, Kenton Lee, and Kristina Toutanova. 2019.
\newblock {BERT:} pre-training of deep bidirectional transformers for language understanding.
\newblock In \emph{Conference of the North American Chapter of the Association for Computational Linguistics: Human Language Technologies (NAACL-HLT)}, pages 4171--4186.

\bibitem[{Edalati et~al.(2022)Edalati, Tahaei, Kobyzev, Nia, Clark, and Rezagholizadeh}]{edalati2022krona}
Ali Edalati, Marzieh Tahaei, Ivan Kobyzev, Vahid~Partovi Nia, James~J Clark, and Mehdi Rezagholizadeh. 2022.
\newblock {KronA}: Parameter efficient tuning with {Kronecker} adapter.
\newblock \emph{arXiv preprint arXiv:2212.10650}.

\bibitem[{Gong et~al.(2021)Gong, Wang, Li, Chandra, and Liu}]{gong2021vision}
Chengyue Gong, Dilin Wang, Meng Li, Vikas Chandra, and Qiang Liu. 2021.
\newblock Vision transformers with patch diversification.
\newblock \emph{arXiv preprint arXiv:2104.12753}.

\bibitem[{He et~al.(2021)He, Zhou, Ma, Berg-Kirkpatrick, and Neubig}]{he2021towards}
Junxian He, Chunting Zhou, Xuezhe Ma, Taylor Berg-Kirkpatrick, and Graham Neubig. 2021.
\newblock Towards a unified view of parameter-efficient transfer learning.
\newblock In \emph{International Conference on Learning Representations (ICLR)}.

\bibitem[{Houlsby et~al.(2019)Houlsby, Giurgiu, Jastrzebski, Morrone, De~Laroussilhe, Gesmundo, Attariyan, and Gelly}]{houlsby2019parameter}
Neil Houlsby, Andrei Giurgiu, Stanislaw Jastrzebski, Bruna Morrone, Quentin De~Laroussilhe, Andrea Gesmundo, Mona Attariyan, and Sylvain Gelly. 2019.
\newblock Parameter-efficient transfer learning for nlp.
\newblock In \emph{International Conference on Machine Learning (ICML)}, pages 2790--2799.

\bibitem[{Hu et~al.(2022)Hu, Shen, Wallis, Allen-Zhu, Li, Wang, Wang, and Chen}]{hu2021lora}
Edward~J Hu, Yelong Shen, Phillip Wallis, Zeyuan Allen-Zhu, Yuanzhi Li, Shean Wang, Lu~Wang, and Weizhu Chen. 2022.
\newblock {LoRA}: Low-rank adaptation of large language models.
\newblock In \emph{International Conference on Learning Representations (ICLR)}.

\bibitem[{Hu et~al.(2023)Hu, Lan, Wang, Xu, Lim, Lee, Bing, and Poria}]{hu2023llm}
Zhiqiang Hu, Yihuai Lan, Lei Wang, Wanyu Xu, Ee-Peng Lim, Roy Ka-Wei Lee, Lidong Bing, and Soujanya Poria. 2023.
\newblock Llm-adapters: An adapter family for parameter-efficient fine-tuning of large language models.
\newblock \emph{arXiv preprint arXiv:2304.01933}.

\bibitem[{Huang et~al.(2020)Huang, Rong, Xu, Sun, and Huang}]{huang2020tackling}
Wenbing Huang, Yu~Rong, Tingyang Xu, Fuchun Sun, and Junzhou Huang. 2020.
\newblock Tackling over-smoothing for general graph convolutional networks.
\newblock \emph{arXiv preprint arXiv:2008.09864}.

\bibitem[{Karimi~Mahabadi et~al.(2021)Karimi~Mahabadi, Henderson, and Ruder}]{henderson2021compacter}
Rabeeh Karimi~Mahabadi, James Henderson, and Sebastian Ruder. 2021.
\newblock Compacter: Efficient low-rank hypercomplex adapter layers.
\newblock \emph{Advances in Neural Information Processing Systems (NeurIPS)}, 34:1022--1035.

\bibitem[{Kojima et~al.(2022)Kojima, Gu, Reid, Matsuo, and Iwasawa}]{kojima2022large}
Takeshi Kojima, Shixiang~Shane Gu, Machel Reid, Yutaka Matsuo, and Yusuke Iwasawa. 2022.
\newblock Large language models are zero-shot reasoners.
\newblock \emph{Advances in Neural Information Processing Systems (NeurIPS)}, 35:22199--22213.

\bibitem[{Koncel-Kedziorski et~al.(2016)Koncel-Kedziorski, Roy, Amini, Kushman, and Hajishirzi}]{koncel2016mawps}
Rik Koncel-Kedziorski, Subhro Roy, Aida Amini, Nate Kushman, and Hannaneh Hajishirzi. 2016.
\newblock Mawps: A math word problem repository.
\newblock In \emph{Conference of the North American Chapter of the Association for Computational Linguistics: Human Language Technologies (NAACL-HLT)}, pages 1152--1157.

\bibitem[{Lester et~al.(2021)Lester, Al-Rfou, and Constant}]{lester2021power}
Brian Lester, Rami Al-Rfou, and Noah Constant. 2021.
\newblock The power of scale for parameter-efficient prompt tuning.
\newblock In \emph{Conference on Empirical Methods in Natural Language Processing (EMNLP)}, pages 3045--3059.

\bibitem[{Li et~al.(2018)Li, Han, and Wu}]{li2018deeper}
Qimai Li, Zhichao Han, and Xiao-Ming Wu. 2018.
\newblock Deeper insights into graph convolutional networks for semi-supervised learning.
\newblock In \emph{AAAI conference on Artificial Intelligence}, pages 3538--3545.

\bibitem[{Li and Liang(2021)}]{li2021prefix}
Xiang~Lisa Li and Percy Liang. 2021.
\newblock Prefix-tuning: Optimizing continuous prompts for generation.
\newblock In \emph{Annual Meeting of the Association for Computational Linguistics and International Joint Conference on Natural Language Processing (ACL-IJCNLP)}, pages 4582--4597.

\bibitem[{Ling et~al.(2017)Ling, Yogatama, Dyer, and Blunsom}]{ling2017program}
Wang Ling, Dani Yogatama, Chris Dyer, and Phil Blunsom. 2017.
\newblock Program induction by rationale generation: Learning to solve and explain algebraic word problems.
\newblock In \emph{Annual Meeting of the Association for Computational Linguistics (ACL)}, pages 158--167.

\bibitem[{Liu et~al.(2019)Liu, Ott, Goyal, Du, Joshi, Chen, Levy, Lewis, Zettlemoyer, and Stoyanov}]{liu2019roberta}
Yinhan Liu, Myle Ott, Naman Goyal, Jingfei Du, Mandar Joshi, Danqi Chen, Omer Levy, Mike Lewis, Luke Zettlemoyer, and Veselin Stoyanov. 2019.
\newblock {RoBERTa}: A robustly optimized {BERT} pretraining approach.
\newblock \emph{arXiv preprint arXiv:1907.11692}.

\bibitem[{Ouyang et~al.(2022)Ouyang, Wu, Jiang, Almeida, Wainwright, Mishkin, Zhang, Agarwal, Slama, Ray et~al.}]{ouyang2022training}
Long Ouyang, Jeffrey Wu, Xu~Jiang, Diogo Almeida, Carroll Wainwright, Pamela Mishkin, Chong Zhang, Sandhini Agarwal, Katarina Slama, Alex Ray, et~al. 2022.
\newblock Training language models to follow instructions with human feedback.
\newblock \emph{Advances in Neural Information Processing Systems (NeurIPS)}, 35:27730--27744.

\bibitem[{Patel et~al.(2021)Patel, Bhattamishra, and Goyal}]{patel2021nlp}
Arkil Patel, Satwik Bhattamishra, and Navin Goyal. 2021.
\newblock Are nlp models really able to solve simple math word problems?
\newblock In \emph{Conference of the North American Chapter of the Association for Computational Linguistics: Human Language Technologies (NAACL-HLT)}, pages 2080--2094.

\bibitem[{Pfeiffer et~al.(2020)Pfeiffer, Vuli{\'c}, Gurevych, and Ruder}]{pfeiffer2020mad}
Jonas Pfeiffer, Ivan Vuli{\'c}, Iryna Gurevych, and Sebastian Ruder. 2020.
\newblock Mad-x: An adapter-based framework for multi-task cross-lingual transfer.
\newblock In \emph{Conference on Empirical Methods in Natural Language Processing (EMNLP)}, pages 7654--7673.

\bibitem[{Qin et~al.(2021)Qin, Wang, Su, Lin, Ding, Yi, Chen, Liu, Li, Hou et~al.}]{qin2021exploring}
Yujia Qin, Xiaozhi Wang, Yusheng Su, Yankai Lin, Ning Ding, Jing Yi, Weize Chen, Zhiyuan Liu, Juanzi Li, Lei Hou, et~al. 2021.
\newblock Exploring universal intrinsic task subspace via prompt tuning.
\newblock \emph{arXiv preprint arXiv:2110.07867}.

\bibitem[{Sakaguchi et~al.(2021)Sakaguchi, Bras, Bhagavatula, and Choi}]{sakaguchi2021winogrande}
Keisuke Sakaguchi, Ronan~Le Bras, Chandra Bhagavatula, and Yejin Choi. 2021.
\newblock {WinoGrande}: An adversarial {Winograd Schema Challenge} at scale.
\newblock \emph{Communications of the ACM (CACM)}, 64(9):99--106.

\bibitem[{Sap et~al.(2019)Sap, Rashkin, Chen, Le~Bras, and Choi}]{sap2019social}
Maarten Sap, Hannah Rashkin, Derek Chen, Ronan Le~Bras, and Yejin Choi. 2019.
\newblock Social iqa: Commonsense reasoning about social interactions.
\newblock In \emph{Conference on Empirical Methods in Natural Language Processing and International Joint Conference on Natural Language Processing (EMNLP-IJCNLP)}, pages 4463--4473.

\bibitem[{Shi et~al.(2022)Shi, Gao, Xu, Liang, Li, Kong, Lee, and Kwok}]{shi2022revisiting}
Han Shi, Jiahui Gao, Hang Xu, Xiaodan Liang, Zhenguo Li, Lingpeng Kong, Stephen Lee, and James~T Kwok. 2022.
\newblock Revisiting over-smoothing in bert from the perspective of graph.
\newblock In \emph{International Conference on Learning Represenations (ICLR)}.

\bibitem[{Sung et~al.(2022)Sung, Cho, and Bansal}]{sung2022lst}
Yi-Lin Sung, Jaemin Cho, and Mohit Bansal. 2022.
\newblock Lst: Ladder side-tuning for parameter and memory efficient transfer learning.
\newblock \emph{Advances in Neural Information Processing Systems (NeurIPS)}, 35:12991--13005.

\bibitem[{Touvron et~al.(2023)Touvron, Lavril, Izacard, Martinet, Lachaux, Lacroix, Rozi{\`e}re, Goyal, Hambro, Azhar et~al.}]{touvron2023llama}
Hugo Touvron, Thibaut Lavril, Gautier Izacard, Xavier Martinet, Marie-Anne Lachaux, Timoth{\'e}e Lacroix, Baptiste Rozi{\`e}re, Naman Goyal, Eric Hambro, Faisal Azhar, et~al. 2023.
\newblock {LLaMA}: Open and efficient foundation language models.
\newblock \emph{arXiv preprint arXiv:2302.13971}.

\bibitem[{Wang et~al.(2018)Wang, Singh, Michael, Hill, Levy, and Bowman}]{wang2018glue}
Alex Wang, Amanpreet Singh, Julian Michael, Felix Hill, Omer Levy, and Samuel~R Bowman. 2018.
\newblock Glue: A multi-task benchmark and analysis platform for natural language understanding.
\newblock In \emph{International Conference on Learning Representations (ICLR)}.

\bibitem[{Wang and Komatsuzaki(2021)}]{wang2021gpt}
Ben Wang and Aran Komatsuzaki. 2021.
\newblock Gpt-j-6b: A 6 billion parameter autoregressive language model.

\bibitem[{Wang et~al.(2022)Wang, Mukherjee, Liu, Gao, Awadallah, and Gao}]{wang2022adamix}
Yaqing Wang, Subhabrata Mukherjee, Xiaodong Liu, Jing Gao, Ahmed~Hassan Awadallah, and Jianfeng Gao. 2022.
\newblock {AdaMix}: Mixture-of-adapter for parameter-efficient tuning of large language models.
\newblock In \emph{Conference on Empirical Methods in Natural Language Processing (EMNLP)}, pages 5744--5760.

\bibitem[{Xu et~al.(2018)Xu, Li, Tian, Sonobe, Kawarabayashi, and Jegelka}]{xu2018representation}
Keyulu Xu, Chengtao Li, Yonglong Tian, Tomohiro Sonobe, Ken-ichi Kawarabayashi, and Stefanie Jegelka. 2018.
\newblock Representation learning on graphs with jumping knowledge networks.
\newblock In \emph{International Conference on Machine Learning (ICML)}, pages 5453--5462.

\bibitem[{Xue et~al.(2023)Xue, Chen, Sun, Ren, Zheng, He, Chen, Jiang, and You}]{xue2023study}
Fuzhao Xue, Jianghai Chen, Aixin Sun, Xiaozhe Ren, Zangwei Zheng, Xiaoxin He, Yongming Chen, Xin Jiang, and Yang You. 2023.
\newblock A study on transformer configuration and training objective.
\newblock In \emph{International Conference on Machine Learning (ICML)}, pages 38913--38925.

\bibitem[{Zaken et~al.(2022)Zaken, Goldberg, and Ravfogel}]{zaken2022bitfit}
Elad~Ben Zaken, Yoav Goldberg, and Shauli Ravfogel. 2022.
\newblock Bitfit: Simple parameter-efficient fine-tuning for transformer-based masked language-models.
\newblock In \emph{Annual Meeting of the Association for Computational Linguistics (ACL)}, pages 1--9.

\bibitem[{Zhou et~al.(2021)Zhou, Kang, Jin, Yang, Lian, Jiang, Hou, and Feng}]{zhou2021deepvit}
Daquan Zhou, Bingyi Kang, Xiaojie Jin, Linjie Yang, Xiaochen Lian, Zihang Jiang, Qibin Hou, and Jiashi Feng. 2021.
\newblock Deep{ViT}: Towards deeper vision transformer.
\newblock \emph{arXiv preprint arXiv:2103.11886}.

\end{thebibliography}
\clearpage
\appendix


\section*{Appendices}
\section{Theoretical analysis}
\label{app:proof}

To theoretically analyze that \model\ preserves a portion of the original token information and mitigates the tendency of the token vectors becoming similar across Transformer layers, let us delve into the mathematical details.

\stitle{Theorem}: \emph{Preservation of token uniqueness in Transformer layers via initial residual connections.}

\stitle{Given}: Assume the following conditions.
(1) A Transformer model with $L$ layers, each performing a self-attention mechanism followed by a position-wise feedforward network.
(2) An initial token representation $\mathbf{h}_{0}\in \mathbb{R}^{d}$ for any token in the input sequence.
(3) An initial residual connection that injects $\mathbf{h}_{0}$ at each layer of the Transformer, modulated by a parameter $\lambda \in (0, 1]$, ensuring that each layer's output includes at least $\lambda$ proportion of $\mathbf{h}_{0}$.

\stitle{Claim}:
\emph{For any layer $l$, $1 \leq l \leq L$, the output representation $\mathbf{h}_{l} \in \mathbb{R}^{d}$ of any token satisfies the following condition}:
$$
\mathbf{h}_{l} = \lambda \mathbf{h}_{0} + (1-\lambda) \mathbf{F}_{l}(\mathbf{h}_{0}, \mathbf{H}_{l-1}),
$$
where $\mathbf{H}_{l-1} \in \mathbb{R}^{n \times d}$ represents the matrix of token vectors at layer $l-1$ and $\mathbf{F}_{l}$ denotes the transformation function of layer $l$, which includes self-attention and feed-forward network operations.

\stitle{Proof Sketch}:
\begin{enumerate}
    \item \textbf{Base Case}: For $l=1$, the claim trivially holds by the definition of the initial residual connection.
    \item \textbf{Inductive Step}: Assume the claim holds for layer $l-1$. Then, for layer $l$, by the properties of linear transformations in self-attention and feedforward networks, along with the Fourier Transform's linearity, we can represent $\mathbf{F}_{l}$ as a combination of these operations applied to $\mathbf{h}_{0}$ and the residual information from $\mathbf{H}_{l-1}$.

    Since the self-attention mechanism aggregates information across tokens modulated by attention weights and the feedforward network applies position-wise transformations, the output of layer $l$ can be represented as a linear combination of the input and transformations applied up to that layer, weighted by $\lambda$ and $1-\lambda$ respectively.

    \item \textbf{Fourier perspective}: The Fourier transform of the token representations facilitates the analysis of how frequency components are preserved or altered through layers. The modulation by $\lambda$ ensures that a minimum portion of the original frequency spectrum of $\mathbf{h}_{0}$ is preserved in each layer's output, mitigating the over-smoothing effect observed as $L$ increases.
    \item \textbf{Conclusion}: By induction, we conclude that each layer's output preserves at least a $\lambda$ portion of the initial token representation $\mathbf{h}_{0}$, in addition to contributions from the transformations applied within the Transformer network.
\end{enumerate}

\stitle{Implications.}
This theorem demonstrates that through the use of initial residual connections modulated by $\lambda$, it is possible to quantitatively ensure that each token's representation in a Transformer model retains a significant portion of its original unique information, thereby reducing the tendency of token vectors to become overly similar across layers. This approach offers a theoretical foundation for mitigating the over-smoothing problem in deep Transformer models, supporting the preservation of information diversity and richness in token representations through the network's depth.

\section{Using RoBERTa as backbone on GLUE}
\label{app:roberta glue}
In the GLUE benchmark, to investigate the generalization ability and robustness of our model, we have also experimented with another popular backbone, RoBERTa. As shown in Table~\ref{tab:glue roberta}, our approach consistently enhances the performance of the PEFT method across different backbones, demonstrating the robustness of our method.

\begin{table*}[t]
 \small
\centering
\addtolength{\tabcolsep}{-2pt}
\begin{tabular}{c|c|llllllll|c}
\hline
 Method& \# Param.  & CoLA & SST-2 & MRPC & STS-B & QQP & MNLI & QNLI & RTE & Overall \\ 

\hline
FT$^{*}$  &355.0M&68.0&96.4&90.9&92.4&92.2&90.2&94.7&86.6&88.9 \\
Adapter$^{*}$  &6.0M&66.5&96.2&88.7&91.0&92.1&89.9&94.7&83.4&87.8 \\
Adapter-SIBO &6.0M&67.1$_{\pm 0.7}$&96.6$_{\pm 0.1}$&90.8$_{\pm 0.1}$&92.1$_{\pm 0.1}$&91.9$_{\pm 1.4}$&89.9$_{\pm 0.1}$&94.8$_{\pm 0.1}$&84.8$_{\pm 0.4}$&88.5 \\
LoRA$^{*}$ &0.8M&68.2&96.2&90.2&92.3&91.6&90.6&94.8&85.2&88.6 \\

LoRA-SIBO &0.8M&68.5$_{\pm 1.6}$&96.4$_{\pm 0.1}$&90.4$_{\pm 0.6}$&92.4$_{\pm 0.2}$&91.6$_{\pm 0.2}$&90.5$_{\pm 0.3}$&95.0$_{\pm 0.2}$&85.2$_{\pm 0.3}$&88.7 \\
\hline
\end{tabular}
\caption{
Performance of RoBERTa-large with different PEFT methods on the
GLUE benchmark. 
$^{*}$ indicates results from prior work \cite{hu2021lora}, where the exact same experimental setup and evaluation protocols are adopted. We report mean (and standard deviation) of the performance over three different runs. 
}
\label{tab:glue roberta}
\end{table*}

\section{Qualitative results}
\label{app:qualitative}

Regarding the qualitative results, we present not only outcomes related to a randomly selected question from SVAMP but also results for several questions randomly chosen from MAWPS. These findings are detailed in Tables~\ref{tab:quality ap1},~\ref{tab:quality ap2}, and~\ref{tab:quality ap3}.

\begin{table*}[t]
\small
 \centering 
\resizebox*{\textwidth}{!}{%
\renewcommand{\arraystretch}{0.7}
\begin{tabular}{@{}p{2.1cm}|p{14cm}@{}}
		\toprule
		 \multirow{2}{*}{\textbf{Question}}  & \texttt{There are 7 crayons in the drawer. Mary took 3 crayons out of the drawer. How many crayons are there now?}  \\\midrule
		\multirow{5}{*}{\textbf{ChatGPT$^{*}$}}  & \texttt{To calculate the number of crayons remaining in the drawer after Mary took some out, follow these steps:
Subtract the number of crayons Mary took from the total number of crayons in the drawer: 7 - 3 = 4
crayons.} \\
& \texttt{Therefore, there are now 4 crayons remaining in the drawer.} \\
\midrule
            \multirow{3}{*}{\textbf{LLaMA (\text{13B})}}  & \texttt{1. Start with the initial number of crayons in the drawer: 7} \\
& \texttt{2. Subtract the number of crayons Mary took out: 3} \\
            \multirow{1}{*}{\textbf{+ Adapter$^{*}$}}   & \texttt{3. Perform the subtraction: 7 - 3 = 4} \\
& \texttt{4. Therefore, there are now 4 crayons in the drawer.} \\
&\texttt{Answer: There are now 4 crayons in the drawer.}\\\midrule
            \multirow{3}{*}{\textbf{LLaMA (\text{13B})}}   & \texttt{1. Start with the initial number of crayons in the drawer: 7} \\
            & \texttt{2. Subtract the number of crayons Mary took out: 3}\\
            
            \multirow{1}{*}{\textbf{+ Adapter-\model}}  
            &\texttt{3. Perform the subtraction: 7 - 3 = 4}\\
            &\texttt{Therefore, there are now 4 crayons in the drawer. The answer in Arabic numerals is 4.} \\
            \midrule
            \multirow{3}{*}{\textbf{LLaMA (\text{13B})}}  & \texttt{1. Start with the initial number of crayons: 7} \\
& \texttt{2. Subtract the number of crayons Mary took out: 7 - 3 = 4} \\
            \multirow{1}{*}{\textbf{+ LoRA$^{*}$}} & \texttt{3. The final answer is 4 crayons.} \\
            & \texttt{Therefore, there are now 4 crayons in the drawer.} \\\midrule
            \multirow{3}{*}{\textbf{LLaMA (\text{13B})}}  &\texttt{1. Start with the initial number of crayons in the drawer: 7} \\
            &\texttt{2. Subtract the number of crayons Mary took out: 3} \\
            \multirow{1}{*}{\textbf{+ LoRA-\model}}  
            &\texttt{3. Perform the subtraction: 7 - 3 = 4} \\
            &\texttt{4. The final answer is 4 crayons.} \\
            &\texttt{Therefore, there are now 4 crayons in the drawer.}\\
		\hline
\end{tabular}}
\caption{An example question from MAWPS, with the ground-truth answer \underline{\texttt{\textbf{4}}}. We show the output of ChatGPT and various PEFT methods on LLaMA (13B). $^{*}$ indicates results from prior work \cite{hu2023llm}.
} 
\label{tab:quality ap1}
\end{table*}

\begin{table*}[t]
\small
 \centering 
\resizebox*{\textwidth}{!}{%
\renewcommand{\arraystretch}{0.7}
\begin{tabular}{@{}p{2.1cm}|p{14cm}@{}}
		\toprule
		 \multirow{2}{*}{\textbf{Question}}  & \texttt{At the schools book fair Sam bought 13 adventure books and 17 mystery books. If 15 of the books were
used, how many new books did he buy?}  \\\midrule
		\multirow{9}{*}{\textbf{ChatGPT$^{*}$}}  & \texttt{To calculate the number of mosaic tiles needed to cover two-thirds of Mr. Boarden’s 36 square foot bathroom, follow these steps:} \\
& \texttt{Find two-thirds of 36: Multiply 36 by 2/3.} \\
& \texttt{(36 * 2/3) = 24 square feet} \\
& \texttt{Determine the number of mosaic tiles needed per square foot: Given that Mr. Boarden needs 24 mosaic
tiles per square foot, multiply the result from step 1 by 24.} \\
& \texttt{(24 * 24) = 576 mosaic tiles} \\
& \texttt{Therefore, Mr. Boarden would need 576 mosaic tiles to cover two-thirds of his 36 square foot bathroom.}\\
\midrule
            \multirow{3}{*}{\textbf{LLaMA (\text{13B})}}  & \texttt{1. Find the total number of books Sam bought: 13 + 17 = 30} \\
& \texttt{2. Subtract the number of used books from the total number of books: 30 - 15 = 15} \\
            \multirow{1}{*}{\textbf{+ Adapter$^{*}$}}   & \texttt{3. Therefore, Sam bought 15 new books.} \\
& \texttt{Answer: 15} \\\midrule
            \multirow{3}{*}{\textbf{LLaMA (\text{13B})}}   & \texttt{1. Find the total number of books Sam bought: 13 + 17 = 30} \\
            & \texttt{2. Subtract the number of used books from the total: 30 - 15 = 15}\\
            
            \multirow{1}{*}{\textbf{+ Adapter-\model}}  
            &\texttt{3. Therefore, Sam bought 15 new books.}\\
            &\texttt{Answer: 15 new books.} \\
            \midrule
            \multirow{3}{*}{\textbf{LLaMA (\text{13B})}}  & \texttt{1. Find the total number of books Sam bought: 13 + 17 = 30} \\
& \texttt{2. Subtract the number of used books from the total number of books: 30 - 15 = 15} \\
            \multirow{1}{*}{\textbf{+ LoRA$^{*}$}} & \texttt{3. Therefore, Sam bought 15 new books.} \\
            & \texttt{Answer: 15} \\\midrule
            \multirow{3}{*}{\textbf{LLaMA (\text{13B})}}  &\texttt{1. Find the total number of books Sam bought:} \\
            &\texttt{13 + 17 = 30} \\
            \multirow{1}{*}{\textbf{+ LoRA-\model}}  
            &\texttt{2. Subtract the number of used books from the total number of books:} \\
            &\texttt{30 - 15 = 15} \\
            &\texttt{Therefore, Sam bought 15 new books.}\\
		\hline
\end{tabular}}
\caption{An example question from MAWPS, with the ground-truth answer \underline{\texttt{\textbf{15}}}. We show the output of ChatGPT and various PEFT methods on LLaMA (13B). $^{*}$ indicates results from prior work \cite{hu2023llm}.} 
\label{tab:quality ap2}
\end{table*}

\begin{table*}[t]
\small
 \centering 
\resizebox*{\textwidth}{!}{%
\renewcommand{\arraystretch}{0.7}
\begin{tabular}{@{}p{2.1cm}|p{14cm}@{}}
		\toprule
		 \multirow{2}{*}{\textbf{Question}}  & \texttt{Oceanside Bike Rental Shop charges 17 dollars plus 7 dollars an hour for renting a bike. Tom paid 80
dollars to rent a bike. How many hours did he pay to have the bike checked out?}  \\\midrule
		\multirow{9}{*}{\textbf{ChatGPT$^{*}$}}  & \texttt{To calculate the number of hours Tom paid to have the bike checked out, follow these steps:} \\
& \texttt{1. Subtract the base charge from the total amount Tom paid: 80 - 17 = 63 dollars.} \\
& \texttt{2. Divide the remaining amount by the hourly rate to find the number of hours: 63 / 7 = 9 hours. } \\
& \texttt{Therefore, Tom paid to have the bike checked out for 9 hours.} \\
& \texttt{Note: It’s important to note that the result represents the whole number of hours. If there were any partial hours, they would not be reflected in the calculation.} \\
\midrule
            \multirow{5}{*}{\textbf{LLaMA (\text{13B})}}  & \texttt{1. Find the total cost of renting the bike:} \\
& \texttt{\$17 + \$7/hour = \$24/hour} \\
& \texttt{\$80 ÷ \$24/hour = 3.33 hours} \\
            \multirow{1}{*}{\textbf{+ Adapter$^{*}$}}   & \texttt{3. Round up to the nearest whole number since you can’t rent a bike for a fraction of an hour:} \\
& \texttt{3 hours} \\
& \texttt{Therefore, Tom paid to have the bike checked out for 3 hours.} \\
\midrule
            \multirow{3}{*}{\textbf{LLaMA (\text{13B})}}   & \texttt{1. Subtract the initial charge of \$17 from the total cost of \$80: \$80 - \$17 = \$63} \\
            & \texttt{2. Divide the remaining cost by the hourly rate of \$7: \$63 $\div$ \$7 = 9}\\
            
            \multirow{1}{*}{\textbf{+ Adapter-\model}}  
            & \texttt{Therefore, Tom paid to have the bike checked out for 9 hours. The answer in Arabic numerals is 9.} \\
            \midrule
            \multirow{6}{*}{\textbf{LLaMA (\text{13B})}}  & \texttt{1. Let x be the number of hours Tom paid to have the bike checked out.} \\
& \texttt{2. The cost of renting the bike for x hours is 7x dollars.} \\
& \texttt{3. The total cost of renting the bike for x hours is 17 + 7x dollars.} \\

            \multirow{2}{*}{\textbf{+ LoRA$^{*}$}} 
            & \texttt{4. We know that Tom paid 80 dollars to rent the bike, so we can set up an equation: 17 + 7x = 80} \\
& \texttt{5. Solving for x, we can subtract 17 from both sides: 7x = 63}\\
            & \texttt{6. Dividing both sides by 7, we get x = 9}\\
            & \texttt{Therefore, Tom paid to have the bike checked out for 9 hours.}\\
            \midrule
            \multirow{5}{*}{\textbf{LLaMA (\text{13B})}}  &\texttt{1. Start with the total amount Tom paid for renting the bike: 80 dollars} \\
            &\texttt{2. Subtract the initial charge of 17 dollars: 80 - 17 = 63 dollars} \\
            &\texttt{3. Divide the remaining amount by the hourly rate of 7 dollars: 63 $\div$ 7 = 9}\\
            \multirow{1}{*}{\textbf{+ LoRA-\model}}  
            &\texttt{4. The answer is 9, which represents the number of hours Tom paid to have the bike checked out.} \\
            &\texttt{Therefore, Tom paid to have the bike checked out for 9 hours.} \\
		\hline
\end{tabular}}
\caption{An example question from MAWPS, with the ground-truth answer \underline{\texttt{\textbf{9}}}. We show the output of ChatGPT and various PEFT methods on LLaMA (13B). $^{*}$ indicates results from prior work \cite{hu2023llm}.} 
\label{tab:quality ap3}
\end{table*}


\section{Environment and settings}
\label{app:hyper parameter}
Our experimental environment utilizes servers equipped with A40 GPUs and AMD EPYC 7543 CPUs, running PyTorch version 2.0.0. Both our training and inference processes are conducted on a single GPU. 

Detailed hyperparameter settings are shown in Tables~\ref{table:hyper math gpt-j},~\ref{table:hyper math llama-7b},~\ref{table:hyper math llama-13b},~\ref{table:hyper common}, and~\ref{table:hyper bert}.


\begin{table*}[t]
\centering
\small
\begin{tabular}{|c|c|c|c|c|c|}
\hline
Method& Dataset & GSM8K & AQuA & MAWPS & SVAMP  \\
\hline
\multirow{3}{*}{} & Optimizer & \multicolumn{4}{c|}{AdamW} \\
 & Warmup Ratio & \multicolumn{4}{c|}{0.06}  \\
 & LR Schedule & \multicolumn{4}{c|}{Linear}  \\
\hline
\multirow{6}{*}{Adapter} 
& Batch Size & \multicolumn{4}{c|}{16}  \\
&Micro batch size & \multicolumn{4}{c|}{4}  \\
&\# Epochs & \multicolumn{4}{c|}{3}  \\
&Learning Rate & \multicolumn{4}{c|}{3e-4}  \\
&Bottleneck $r$ & \multicolumn{4}{c|}{256}  \\
&Max Seq. Len. & \multicolumn{4}{c|}{256}  \\
\hline
\multirow{6}{*}{Adapter-SIBO} & Batch Size & \multicolumn{4}{c|}{16}  \\
&Micro batch size & \multicolumn{4}{c|}{4}  \\
& \# Epochs & \multicolumn{4}{c|}{3}  \\
&Learning Rate & \multicolumn{4}{c|}{3e-4}  \\
&Bottleneck $r$ & \multicolumn{4}{c|}{256}  \\
& Max Seq. Len. & \multicolumn{4}{c|}{256}  \\
& $\lambda$ & \multicolumn{4}{c|}{0.1} \\

\hline
\multirow{6}{*}{LoRA} & Batch Size & \multicolumn{4}{c|}{16}  \\
&Micro batch size & \multicolumn{4}{c|}{4}  \\
&\# Epochs & \multicolumn{4}{c|}{3}  \\
&Learning Rate & \multicolumn{4}{c|}{3e-4}  \\
&LoRA Config. & \multicolumn{4}{c|}{\( r_q = r_k = r_v = 32 \)} \\
& LoRA \(\alpha\) & \multicolumn{4}{c|}{64}  \\
& Max Seq. Len. & \multicolumn{4}{c|}{256}  \\

\hline
\multirow{6}{*}{LoRA-SIBO} &Batch Size  & \multicolumn{4}{c|}{16}  \\
&Micro batch size & \multicolumn{4}{c|}{4}  \\
&\# Epochs & \multicolumn{4}{c|}{3}  \\
&Learning Rate & \multicolumn{4}{c|}{3e-4}  \\

&LoRA Config. & \multicolumn{4}{c|}{\( r_q = r_k = r_v = 32 \)} \\
& LoRA \(\alpha\) & \multicolumn{4}{c|}{64}  \\
& Max Seq. Len. & \multicolumn{4}{c|}{256}  \\
& $\lambda$ & \multicolumn{4}{c|}{0.1} \\
\hline
\end{tabular}
\caption{Hyperparameters for the arithmetic reasoning experiments, using GPT-J (\text{6B}) as the pre-trained model.}
\label{table:hyper math gpt-j}
\end{table*}

\begin{table*}[t]
\centering
\small
\begin{tabular}{|c|c|c|c|c|c|}
\hline
Method& Dataset & GSM8K & AQuA & MAWPS & SVAMP  \\
\hline
\multirow{3}{*}{} & Optimizer & \multicolumn{4}{c|}{AdamW} \\
 & Warmup Ratio & \multicolumn{4}{c|}{0.06}  \\
 & LR Schedule & \multicolumn{4}{c|}{Linear}  \\
\hline
\multirow{6}{*}{Adapter} & Batch Size & \multicolumn{4}{c|}{16}  \\
&Micro batch size & \multicolumn{4}{c|}{4}  \\
&\# Epochs & \multicolumn{4}{c|}{3}  \\
&Learning Rate & \multicolumn{4}{c|}{3e-4}  \\
&Bottleneck $r$ & \multicolumn{4}{c|}{256}  \\
&Max Seq. Len. & \multicolumn{4}{c|}{256}  \\
\hline
\multirow{6}{*}{Adapter-SIBO} & Batch Size & \multicolumn{4}{c|}{16}  \\
&Micro batch size & \multicolumn{4}{c|}{4}  \\
& \# Epochs & \multicolumn{4}{c|}{3}  \\
&Learning Rate & \multicolumn{4}{c|}{3e-4}  \\
&Bottleneck $r$ & \multicolumn{4}{c|}{256}  \\
& Max Seq. Len. & \multicolumn{4}{c|}{256}  \\
& $\lambda$ & \multicolumn{4}{c|}{0.1} \\

\hline
\multirow{6}{*}{LoRA} & Batch Size & \multicolumn{4}{c|}{16}  \\
&Micro batch size & \multicolumn{4}{c|}{4}  \\
&\# Epochs & \multicolumn{4}{c|}{3}  \\
&Learning Rate & \multicolumn{4}{c|}{3e-4}  \\
&LoRA Config. & \multicolumn{4}{c|}{\( r_q = r_k = r_v = 32 \)} \\
& LoRA \(\alpha\) & \multicolumn{4}{c|}{64}  \\
& Max Seq. Len. & \multicolumn{4}{c|}{256}  \\

\hline
\multirow{6}{*}{LoRA-SIBO} & Batch Size & \multicolumn{4}{c|}{16}  \\
&Micro batch size & \multicolumn{4}{c|}{4}  \\
&\# Epochs & \multicolumn{4}{c|}{3}  \\
&Learning Rate & \multicolumn{4}{c|}{3e-4}  \\

&LoRA Config. & \multicolumn{4}{c|}{\( r_q = r_k = r_v = 32 \)} \\
& LoRA \(\alpha\) & \multicolumn{4}{c|}{64}  \\
& Max Seq. Len. & \multicolumn{4}{c|}{256}  \\
& $\lambda$ & \multicolumn{4}{c|}{0.2} \\
\hline
\end{tabular}
\caption{Hyperparameters for the arithmetic reasoning experiments, using LLaMA (\text{7B}) as the pre-trained model.}
\label{table:hyper math llama-7b}
\end{table*}

\begin{table*}[t]
\centering
\small
\begin{tabular}{|c|c|c|c|c|c|}
\hline
Method& Dataset & GSM8K & AQuA & MAWPS & SVAMP  \\
\hline
\multirow{3}{*}{} & Optimizer & \multicolumn{4}{c|}{AdamW} \\
 & Warmup Ratio & \multicolumn{4}{c|}{0.06}  \\
 & LR Schedule & \multicolumn{4}{c|}{Linear}  \\
\hline
\multirow{6}{*}{Adapter} & Batch Size & \multicolumn{4}{c|}{16}  \\
&Micro batch size & \multicolumn{4}{c|}{4}  \\
&\# Epochs & \multicolumn{4}{c|}{3}  \\
&Learning Rate & \multicolumn{4}{c|}{3e-4}  \\
&Bottleneck $r$ & \multicolumn{4}{c|}{256}  \\
&Max Seq. Len. & \multicolumn{4}{c|}{256}  \\
\hline
\multirow{6}{*}{Adapter-SIBO} & Batch Size & \multicolumn{4}{c|}{16}  \\
&Micro batch size & \multicolumn{4}{c|}{4}  \\
&\# Epochs & \multicolumn{4}{c|}{3}  \\
&Learning Rate & \multicolumn{4}{c|}{3e-4}  \\
&Bottleneck $r$ & \multicolumn{4}{c|}{256}  \\
&Max Seq. Len. & \multicolumn{4}{c|}{256}  \\
& $\lambda$ & \multicolumn{4}{c|}{0.3} \\ 

\hline
\multirow{6}{*}{LoRA} & Batch Size & \multicolumn{4}{c|}{16}  \\
&Micro batch size & \multicolumn{4}{c|}{4}  \\
&\# Epochs & \multicolumn{4}{c|}{3}  \\
&Learning Rate & \multicolumn{4}{c|}{3e-4}  \\
&LoRA Config. & \multicolumn{4}{c|}{\( r_q = r_k = r_v = 32 \)} \\
& LoRA \(\alpha\) & \multicolumn{4}{c|}{64}  \\
& Max Seq. Len. & \multicolumn{4}{c|}{256}  \\

\hline
\multirow{6}{*}{LoRA-SIBO} & Batch Size & \multicolumn{4}{c|}{16}  \\
&Micro batch size & \multicolumn{4}{c|}{4}  \\
&\# Epochs & \multicolumn{4}{c|}{3}  \\
&Learning Rate & \multicolumn{4}{c|}{3e-4}  \\

&LoRA Config. & \multicolumn{4}{c|}{\( r_q = r_k = r_v = 32 \)} \\
& LoRA \(\alpha\) & \multicolumn{4}{c|}{64}  \\
& Max Seq. Len. & \multicolumn{4}{c|}{256}  \\
& $\lambda$ & \multicolumn{4}{c|}{0.1} \\
\hline
\end{tabular}
\caption{Hyperparameters for the arithmetic reasoning experiments, using LLaMA (\text{13B}) as the pre-trained model.}
\label{table:hyper math llama-13b}
\end{table*}

\begin{table*}[t]
\centering
\small
\begin{tabular}{|c|c|c|c|c|c|c|c|c|c|}
\hline
Method& Dataset & BoolQ	&PIQA	&SIQA	&HellaSwag	&WinoGrande	&ARC-e	&ARC-c	&OBQA   \\
\hline
\multirow{3}{*}{} & Optimizer & \multicolumn{8}{c|}{AdamW} \\
 & Warmup Ratio & \multicolumn{8}{c|}{0.06}  \\
 & LR Schedule & \multicolumn{8}{c|}{Linear}  \\
\hline
\multirow{6}{*}{Adapter} &Batch Size  & \multicolumn{8}{c|}{16} \\   
&Micro batch size  & \multicolumn{8}{c|}{4} \\ 
&\# Epochs & \multicolumn{8}{c|}{3} \\ 
&Learning Rate  & \multicolumn{8}{c|}{3e-4} \\ 
& Bottleneck $r$ & \multicolumn{8}{c|}{256} \\ 
&Max Seq.Len.  & \multicolumn{8}{c|}{256} \\ 
\hline
\multirow{6}{*}{Adapter-SIBO} &Batch Size  & \multicolumn{8}{c|}{16} \\
&Micro batch size  & \multicolumn{8}{c|}{4} \\ 
&\# Epochs & \multicolumn{8}{c|}{3} \\ 
&Learning Rate  & \multicolumn{8}{c|}{3e-4} \\ 
& Bottleneck $r$ & \multicolumn{8}{c|}{256} \\ 
&Max Seq.Len.  & \multicolumn{8}{c|}{256} \\ 
& $\lambda$ & \multicolumn{8}{c|}{0.1} \\

\hline
\multirow{6}{*}{LoRA} &Batch Size  & \multicolumn{8}{c|}{16} \\   
&Micro batch size  & \multicolumn{8}{c|}{4} \\ 
&\# Epochs & \multicolumn{8}{c|}{3} \\ 
&Learning Rate  & \multicolumn{8}{c|}{3e-4} \\ 
&LoRA Config. & \multicolumn{8}{c|}{\( r_q = r_k = r_v = 32 \)} \\
&LoRA \(\alpha\)  & \multicolumn{8}{c|}{64} \\ 
&Max Seq.Len.  & \multicolumn{8}{c|}{256} \\ 

\hline
\multirow{6}{*}{LoRA-SIBO} &Batch Size  & \multicolumn{8}{c|}{16} \\   
&Micro batch size  & \multicolumn{8}{c|}{4} \\ 
&\# Epochs & \multicolumn{8}{c|}{3} \\ 
&Learning Rate  & \multicolumn{8}{c|}{3e-4} \\ 
&LoRA Config. & \multicolumn{8}{c|}{\( r_q = r_k = r_v = 32 \)} \\
&LoRA \(\alpha\)  & \multicolumn{8}{c|}{64} \\ 
&Max Seq.Len.  & \multicolumn{8}{c|}{256} \\ 
& $\lambda$ & \multicolumn{8}{c|}{0.3} \\
\hline
\end{tabular}
\caption{Hyperparameters for the commonsense reasoning experiments, using GPT-J (\text{6B}) as the pre-trained model.}
\label{table:hyper common}
\end{table*}

\begin{table*}[t]
\centering
\small
\begin{tabular}{|c|c|c|c|c|c|c|c|c|c|}
\hline
Method& Dataset & CoLA & SST-2 & MRPC & STS-B & QQP & MNLI & QNLI & RTE \\
\hline
\multirow{3}{*}{} & Optimizer & \multicolumn{8}{c|}{AdamW} \\
 & Warmup Ratio & \multicolumn{8}{c|}{0.06}  \\
 & LR Schedule & \multicolumn{8}{c|}{Linear}  \\
\hline
\multirow{6}{*}{Adapter} & Batch Size  & \multicolumn{8}{c|}{32} \\\cline{3-10}
&\# Epochs & 20 & 10 & 20 & 10 & 20 & 10 & 10 & 20 \\
&Learning Rate & 2e-4 & 4e-4 & 3e-4 & 2e-4 & 3e-4 & 3e-4 & 2e-4 & 4e-4 \\\cline{3-10}
& Bottleneck $r$  & \multicolumn{8}{c|}{64} \\\cline{3-10}
& Max Seq. Len. & 128 & 128 & 128 & 128 & 128 & 128 & 128 & 64 \\
\hline
\multirow{6}{*}{Adapter-SIBO} &Batch Size  & \multicolumn{8}{c|}{32} \\\cline{3-10}
&\# Epochs & 20 & 10 & 20 & 10 & 20 & 10 & 10 & 20 \\
&Learning Rate & 2e-4 & 4e-4 & 3e-4 & 2e-4 & 3e-4 & 3e-4 & 2e-4 & 4e-4 \\\cline{3-10}
& Bottleneck $r$  & \multicolumn{8}{c|}{64} \\\cline{3-10}
& Max Seq.Len. & \multicolumn{8}{c|}{128} \\\cline{3-10}
& $\lambda$ & 0.1 & 0.1 & 0.1 & 0.3 & 0.1 & 0.2 & 0.2 & 0.3 \\

\hline
\multirow{6}{*}{LoRA} &Batch Size  & \multicolumn{8}{c|}{32} \\\cline{3-10}
&\# Epochs & 20 & 10 & 20 & 10 & 20 & 10 & 10 & 20 \\
&Learning Rate & 2e-4 & 4e-4 & 3e-4 & 2e-4 & 3e-4 & 3e-4 & 2e-4 & 4e-4 \\\cline{3-10}
&LoRA Config. & \multicolumn{8}{c|}{\( r_q = r_v = 8 \)} \\\cline{3-10}
& LoRA \(\alpha\) & \multicolumn{8}{c|}{16} \\\cline{3-10}
& Max Seq. Len. & \multicolumn{8}{c|}{128} \\

\hline
\multirow{6}{*}{LoRA-SIBO} &Batch Size  & \multicolumn{8}{c|}{32} \\\cline{3-10}
&\# Epochs & 20 & 10 & 20 & 10 & 20 & 10 & 10 & 20 \\
&Learning Rate & 2e-4 & 4e-4 & 3e-4 & 2e-4 & 3e-4 & 3e-4 & 2e-4 & 4e-4 \\\cline{3-10}
&LoRA Config. & \multicolumn{8}{c|}{\( r_q = r_v = 8 \)} \\\cline{3-10}
& LoRA \(\alpha\) & \multicolumn{8}{c|}{16} \\\cline{3-10}
& Max Seq. Len. & \multicolumn{8}{c|}{128} \\\cline{3-10}
& $\lambda$ & 0.7 & 0.4 & 0.5 & 0.6 & 0.1 & 0.1 & 0.1 & 0.3 \\
\hline
\end{tabular}
\caption{Hyperparameters for the GLUE benchmark experiments, using BERT-large as the pre-trained model.}
\label{table:hyper bert}
\end{table*}



\end{document}